\documentclass[10pt,twocolumn,letterpaper]{article}

\usepackage[pagenumbers]{cvpr} %

\usepackage{multirow}
\usepackage{makecell}
\newcommand{\std}[1]{{\scriptsize$\pm$#1}}   %
\newcommand{\method}{TesseraFlow}
\newcommand{\invivo}{\emph{in vivo}}   %
\newcommand{\Invivo}{\emph{In vivo}}
\newcommand{\sampler}{ReTile}   %
\usepackage{newfloat}           %
\DeclareFloatingEnvironment[fileext=loa,name=Algorithm,placement=t]{algorithm}

\makeatletter
\g@addto@macro\normalsize{%
  \setlength\abovedisplayskip{5pt plus 2pt minus 3pt}%
  \setlength\belowdisplayskip{5pt plus 2pt minus 3pt}}
\makeatother
\newcommand{\reviewfill}{\iftoggle{cvprfinal}{}{\looseness=1}}

\newcommand{\irbname}{the authors' institutional review board (name withheld for anonymous review)}
\newcommand{\videoonly}[1]{#1}   %

\renewcommand{\irbname}{the institutional review board}
\renewcommand{\videoonly}[1]{}
\definecolor{cvprblue}{rgb}{0.21,0.49,0.74}
\usepackage[pagebackref,breaklinks,colorlinks,allcolors=cvprblue]{hyperref}

\def\confName{CVPR}
\def\confYear{2027}

\title{TesseraFlow: Scalable Video Diffusion \\ for Dynamic Medical Imaging Inverse Problems}

\author{
Atakan Topcu$^{1}$ \qquad
Mohammadhossein Momeni$^{1}$ \qquad 
Alfredo De Goyeneche$^{1}$ \qquad
Michael Lustig$^{1}$\\
$^{1}$Electrical Engineering and Computer Sciences, University of California, Berkeley\\
{\tt\small \{
atakan\_topcu,\;
hosseinm,\;
asdegoyeneche,\;
mikilustig\}@berkeley.edu }
}

\begin{document}
\addtocontents{toc}{\protect\setcounter{tocdepth}{-10}}%
\twocolumn[{%
\renewcommand\twocolumn[1][]{#1}%
\maketitle
\begin{center}
    \centering
    \captionsetup{type=figure}
    \includegraphics[width=0.99\textwidth]{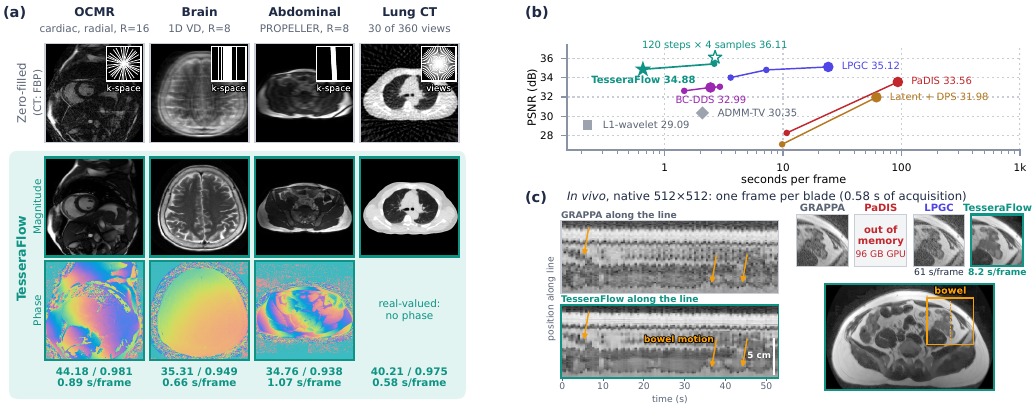}
    \captionof{figure}{\textbf{\method{} can reconstruct dynamic medical images with a generative prior at about one frame per second, an order of magnitude faster than patch and latent priors.} (a)~One test frame per dataset at full field of view: the zero-filled reconstruction (CT: filtered backprojection) with its k-space sampling (inset), and the magnitude and phase of \method{} (teal; CT is real-valued), with PSNR (dB) / SSIM and time per frame. (b)~Quality against time on brain MRI (1D VD, $R{=}8$) at 60 and 120 network evaluations and as published (large dot). PaDIS and Latent + DPS fail at 60. (c)~\Invivo{} scan at native $512\times512$: a frame reconstructed with \method{}, and each blade with GRAPPA and \method{} over all 92 frames. The strips show bowel peristalsis, the wave-like contraction of the bowel.}
   
    \label{fig:teaser}
\end{center}

}]
\begin{abstract}
Generative priors are state-of-the-art for modeling medical images in image reconstruction tasks, but reconstruction with video priors takes tens of seconds to minutes per frame. 
Latent priors inherit the errors of their autoencoder, the memory of pixel-space whole-frame priors grows with the frame size, and pixel-space patch priors are slow for two reasons.
First, their deep networks process every frame at full resolution at every sampling step, even though early steps resolve only low frequencies. Second, their samplers spend many network evaluations per sampling step on overlapping windows, multiple re-noising rounds, or backpropagation through the prior. 
We introduce \method{}, a complex-valued pixel-space flow prior that reduces the first cost. 
A global encoder reads a low-resolution view of every frame once per sampling step. A single shallow decoder operates on this view during early steps and on full-resolution space-time tiles during later steps, so no network processes a whole frame at full resolution. 
Our sampler, \sampler{}, addresses the second reason. 
We use randomized grid shifting at every sampling step along with re-noising, which acts as a stochastic gradient instead of computing overlapping tiles, so a single network pass per sampling step suffices. 
On its own, \sampler{} speeds up an existing global-context patch prior by 6--9$\times$, and our prior adds a further 3--5$\times$ speedup. 
\method{} reconstructs about one $256\times256$ frame per second, 20--367$\times$ faster than patch and latent video priors, and can match or exceed the best generative baseline on cardiac, brain, and abdominal MRI and lung CT, at a fraction of its GPU time. Code is available at \href{https://github.com/mikgroup/TesseraFlow}{mikgroup/TesseraFlow}

\begin{figure*}[!t]
    \centering
    \includegraphics[width=0.95\textwidth]{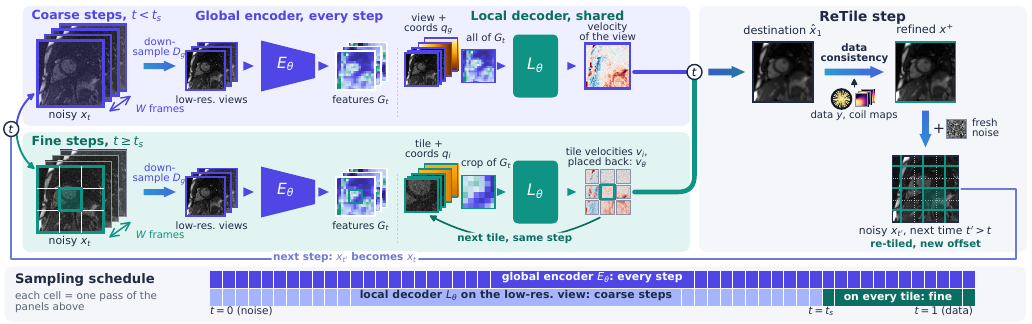}
    \caption[\method{} and \sampler{}]{\textbf{\method{} and \sampler{}.} One sampling step: a coarse step ($t<t_s$, top, purple) or a fine step ($t\geq t_s$, middle, teal), and the schedule of steps (bottom). At every step, the encoder $E_\theta$ reads a low-resolution version of every frame, the size of a tile. In a coarse step, the shallow decoder $L_\theta$ predicts the velocity of this low-resolution video from all encoder features $G_t$. In a fine step, it predicts the velocity of each space-time tile from the tile and the features at its location, and the tile predictions are placed side by side. \sampler{} refines the predicted clean video against the measurements, re-noises it, and moves the tiles to a new random offset for the next step.}
    \label{fig:architecture}
\end{figure*}%

\end{abstract}
    
\section{Introduction}
\label{sec:intro}
Dynamic imaging records anatomy and function over time. Cine MRI follows the beating heart, fetal MRI captures fetal and placental motion, time-resolved abdominal imaging could visualize bowel peristalsis, and 4D CT resolves the breathing cycle. 
 Resolving motion that does not repeat requires reconstructing each frame from far fewer measurements than a static image needs, and the missing information must come from a prior. Cine MRI and 4D CT instead pool measurements over many cardiac or respiratory cycles by gating or binning~\cite{feng2016xdgrasp,keall2004fourdct}, which assumes that the motion repeats and averages out motion that does not, such as irregular breathing. Real-time methods avoid gating with hand-crafted spatio-temporal regularizers such as sparsity and low rank~\cite{lustig2007sparse,otazo2015ls}, but each organ moves and deforms with its own statistics, and such regularizers hold only at moderate undersampling per frame. Learned priors instead model anatomy and motion from data. Generative diffusion and flow priors are among the most expressive of them~\cite{jalal2021robust,chung2022score,chung2023dps,chung2024dds,pourya2025flower}, but they were developed mostly for single images and are expensive for videos. A reconstruction takes tens to hundreds of \emph{sampling steps} from noise to the image, each evaluating the prior on every frame. MRI adds a further requirement. Its measurements are samples of k-space, the spatial frequency domain of a complex-valued image, whose phase arises from both the object and the acquisition system. A prior must therefore also model the phase to be combined exactly with the forward operator.

Existing priors for high-resolution dynamic images each have a limitation (\cref{sec:related}). Latent priors~\cite{rombach2022ldm,varma2025medvae} model a compressed latent code. Every reconstruction inherits the errors of the autoencoder, and data consistency must pass through its decoder, which adds a decoder pass and its gradient to every sampling step. Patch-based pixel priors~\cite{wang2023patch,hu2024padis,lei2026cinediff} operate at the native resolution but see only a fraction of the field of view, and their samplers spend many network evaluations per sampling step on overlapping windows, repeated re-noising, or gradients through the prior. Full-field video priors~\cite{ho2022video,qiu2024stdiff,zhang2026ddimo,furnrohr2026pddr} see everything, but each prior network evaluation takes whole frames at full resolution. Their memory grows with the frame size, and with the video length.

These costs in time and memory limit what can be imaged. For example, consider a minute-long undersampled dynamic abdominal MRI scan at $\approx 0.6$\,s per frame and a $512\times512$ image matrix. Existing patch-based priors take a minute to reconstruct each frame or exceed a 96\,GB GPU (\cref{fig:teaser}c).

Existing priors treat global and local structure alike. Yet the information that ties a region to the rest of the video, such as anatomy, contrast, and motion state, is often low-frequency and inexpensive to represent, whereas full-resolution local detail is not.
\method{} models the two separately without an autoencoder (\cref{fig:architecture}). A global encoder reads a low-resolution view of every frame once per sampling step and produces global features that all tiles share. A shallow local decoder predicts the velocity of the generative flow from these features.  Diffusion and flow sampling resolve low frequencies first~\cite{falck2025fourier}, so in the early, coarse steps the decoder runs only on the low-resolution views of all frames, and only in the late steps does it run on full-resolution space-time tiles. No network ever processes a whole frame at full resolution. Unlike the patch prior LPGC~\cite{yang2025ctpatch}, which recomputes its global features for each patch evaluation, \method{} computes the global features once per sampling step for all tiles and uses one decoder for both its coarse and fine steps.

Our sampler \sampler{} evaluates the prior on one non-overlapping space-time tiling per sampling step. At every step, the tile grid moves to a new random offset, a stochastic estimate of the average over all offsets~\cite{coifman1995translation,figueiredo2003em}, and the refined video is re-noised with fresh noise. The shift moves the seams, and the fresh noise keeps each step's errors from carrying into the next. Each step predicts the clean video, corrects it by conjugate gradients, and re-noises it~\cite{pourya2025flower}, with no gradient through the prior, unlike DPS~\cite{chung2023dps}.

Separating global and local structure makes each evaluation of \method{} efficient, and \sampler{} needs only one evaluation per sampling step. Together, they retain the accuracy of existing pixel-space video priors while removing most of their computational cost (\cref{fig:teaser}b). With the same sampler, number of sampling steps, and GPU, \method{} is 3--5$\times$ faster than LPGC at similar PSNR on cardiac, brain, and abdominal MRI and on lung CT (\cref{tab:prior-sampler}).  Compared with patch and latent video priors run with their published samplers, it reconstructs about one $256\times256$ frame per second, 20--367$\times$ faster than these priors (\cref{fig:teaser}). In a slower setting, with more sampling steps and averaging over several samples, \method{} matches or exceeds the best generative baseline on each dataset while still using a fraction of that baseline's GPU time. 

In summary, our contributions are:
\begin{itemize}
    \item \method, a complex-valued pixel-space flow prior in which no network needs a full-resolution frame. One global encoding per sampling step serves all tiles, and the early, coarse steps run the same decoder on a low-resolution view of each frame.
    
    \item \sampler{}, a posterior sampler with which we show that one randomly offset space-time tiling per sampling step suffices when the video is re-noised with fresh noise, and that network evaluations are better spent on sampling steps than on the overlapping windows or repeated re-noising of patch samplers. It makes an existing global-context patch prior 6--9$\times$ faster than its own sampler.

    \item Experiments on cardiac, brain, and abdominal MRI, lung CT, and an \invivo{} abdominal scan from which \method{} reconstructs a $512\times512$ video on a 0.74\,mm grid with one frame every 0.58\,s.
\end{itemize}

\section{Related Work}
\label{sec:related}

\paragraph{Posterior sampling with generative priors.}
Posterior samplers differ in how they bring in the measurements.
Diffusion posterior sampling (DPS)~\cite{chung2023dps} nudges every denoising step toward the measurements with a gradient through the network, which works for any differentiable forward model but at least doubles the cost of a step, and FlowDPS~\cite{kim2025flowdps} brings this guidance to flows.
Decomposed diffusion sampling (DDS)~\cite{chung2024dds}, decoupled noise annealing~\cite{zhang2025daps}, and split-Gibbs samplers~\cite{wu2024pnpdm,flowsgs2026} instead alternate a prior step with a separate data-consistency step, and Flower~\cite{pourya2025flower} does so for flow models with fresh noise at every step. DiffusionMBIR~\cite{chung2023diffusionmbir} and batch-consistent DDS~\cite{kwon2025video} extend 2D priors to volumes and videos, coupled by total variation along the third axis or by noise shared by all frames. All of them pass the whole image or video through the network at every sampling step. \sampler{} keeps the data-consistency step of Flower, but each of its network evaluations takes one space-time tile or the low-resolution view.

\paragraph{Latent priors.}
Latent priors save computation by modeling a compressed code instead of the image.
Latent diffusion~\cite{rombach2022ldm} trains the generative model on the codes of a pretrained autoencoder, and MedVAE~\cite{varma2025medvae} provides such autoencoders for medical images.
The measurements constrain the image and not the code, so data consistency passes through the decoder, by guidance~\cite{rout2023psld} or explicit consistency steps~\cite{song2024resample}, and the reconstruction can be no more accurate than what the decoder reproduces.
For dynamic MRI, STeP~\cite{zhang2025step} samples a latent video prior with decoupled noise annealing~\cite{zhang2025daps}. Each of its likelihood updates passes through the decoder, and a video takes minutes.
\method{} instead keeps the full-resolution complex image as its state and saves computation inside the network rather than by compressing the signal. Our latent baseline's autoencoder round trip alone reaches only 33.1--38.7\,dB.

\paragraph{Patch priors and global context.}
Patch priors keep pixel space and the exact forward operator, and each network evaluation needs memory only for one patch, but each evaluation also sees only that patch.
Patch Diffusion~\cite{wang2023patch} conditions each crop on its position, PaDIS~\cite{hu2024padis} combines position-aware patch predictions into a prior for the whole image, and PaDIS-MRI~\cite{padismri2025} applies it to complex-valued MRI.
The pieces are made consistent at reconstruction time with many steps and either overlapping crops or a non-overlapping tiling shifted by a new random offset at every iteration, as in PaDIS and DiffusionBlend~\cite{song2024diffusionblend} and, earlier, in wavelet-based restoration~\cite{figueiredo2003em}.
For cardiac cine, CineDiff~\cite{lei2026cinediff} blends overlapping $64\times64\times8$ space-time patches under DPS guidance, at minutes per series.
LPGC~\cite{yang2025ctpatch} adds global context by giving every patch evaluation a downsampled copy of the whole CT volume, recomputed at each evaluation, and averages two randomly offset 3D partitions per sampling step, with fresh noise in between.
In \method{}, the global features are computed once per sampling step for all tiles, and \sampler{} needs one randomly offset space-time tiling per sampling step.

\paragraph{Efficient pixel-space generation.}
Pixel-space generators save computation by separating coarse content from full-resolution detail, within the network or across the sampling steps.
DeCo~\cite{ma2026deco} and PixelDiT~\cite{yu2026pixeldit,yu2026pixeldit2} add a lightweight full-resolution pathway to a coarse model but run on the whole image at every step.
Progressive-resolution generators~\cite{jing2022subspace,jin2025pyramidal,xiao2026spectral} run their early steps on a lower-resolution image, and hierarchical patch diffusion~\cite{skorokhodov2024hpdm} generates video patches conditioned on coarser levels.
\method{} runs its decoder on space-time tiles instead of whole full-resolution frames, and its coarse steps run the same decoder on a low-resolution view while the sample, data consistency, and re-noising stay at full resolution (\cref{sec:sampler}).

\paragraph{Dynamic MRI and CT Reconstruction.}
Non-generative reconstruction combines the measurement model with a hand-crafted regularizer~\cite{ravishankar2020review} or relies on supervised training.
For MRI, the model includes coil sensitivities~\cite{uecker2014espirit} and the regularizer is sparsity or low rank in space and time~\cite{lustig2007sparse,otazo2015ls}. For dynamic CT, it is total variation in space and time~\cite{ritschl2012fourd} or sparsity relative to a prior image reconstructed from all frames~\cite{chen2008piccs}.
Real-time MRI uses these regularizers without gating, but they tolerate only moderate undersampling per frame (\cref{tab:main}).
Supervised unrolled networks~\cite{aggarwal2019modl,sriram2020varnet} are fast but trained for one acquisition.
\method{} learns its prior once and serves different acquisitions; our brain model runs unchanged from $R=8$ to 16 (supplementary material).

\section{Method}
\label{sec:method}
\subsection{Measurement model and flow prior}
We reconstruct a dynamic image, or video, $x\in\mathbb{C}^{T\times N_y\times N_x}$ of $T$ frames from measurements
\begin{equation}
    y=Ax+n,\qquad n\sim\mathcal{N}(0,\sigma^2 I),
    \label{eq:measurement}
\end{equation}
where $n$ is Gaussian noise (complex for MRI) with standard deviation $\sigma$, $I$ is the identity, and $A$ acts frame by frame, as $A_\tau$ on frame $\tau$. In MRI, $y$ holds samples of k-space, taken one after another along a trajectory: Cartesian (line by line), radial (spokes through the center), or PROPELLER~\cite{pipe1999propeller} (rotated bands of lines, called blades). The signal is received simultaneously by several coils, each weighting the image by its spatial sensitivity. For Cartesian sampling, $A_\tau=M_\tau F S$ applies these coil sensitivities $S$, the Fourier transform $F$, and a frame-dependent undersampling mask $M_\tau$; radial and PROPELLER sampling use a nonuniform Fourier transform instead of $F$. For CT, $A_\tau=M_\tau R$ applies the parallel-beam Radon transform $R$, which integrates the image along lines at each view angle, and keeps a frame-dependent set of views $M_\tau$.
\paragraph{Complex-valued modeling.}
MR images are complex valued. \method{} models the real and imaginary parts as two channels, so the prior generates the complex image that data consistency compares with the measurements. The spatial variation of the phase is physical, but its constant offset is not, since the coil sensitivities are defined only up to a common phase. We therefore augment the training data with a random global phase $e^{i\theta}$, $\theta\sim\mathcal{U}[0,2\pi)$, so that the prior learns the local phase structure but not the arbitrary offset~\cite{virtue2017complex}. CT images are real and use one channel.

The prior is a rectified flow~\cite{liu2023rectified,lipman2023flow}, a straight-path diffusion model. For clean data $x_1$ and Gaussian noise $\epsilon$, it is trained on
\begin{equation}
    x_t=(1-t)\epsilon+t x_1,\qquad v^\star=x_1-\epsilon,\qquad t\sim\mathcal{U}(0,1),
    \label{eq:path}
\end{equation}
and the network predicts the flow velocity $v_\theta(x_t,t)$, its estimate of the target $v^\star$.

\subsection{Shared global features and local flow velocity}
For a tile size $p$ (\cref{sec:tessellation}), let $D_g$ be an image-resizing operator that keeps the central $p\times p$ Fourier components of each frame of $x_t$, giving a $p\times p$ image. The global encoder $E_\theta$ maps this low-frequency view of the whole field of view to $G_t$, a low-resolution feature map of each frame:
\begin{equation}
    G_t=E_\theta(D_g x_t,t)\in\mathbb{R}^{C\times W\times \frac{p}{4}\times\frac{p}{4}}.
    \label{eq:global}
\end{equation}
Here $C$ is the number of feature channels, $W$ is the number of frames in a temporal block (\cref{sec:tessellation}), and the encoder mixes information across them. All tiles of a block share this $G_t$. For tile $i$, with extraction operator $P_i$ and normalized coordinates $q_i$, which encode the position of the tile, the local decoder $L_\theta$ predicts the velocity
\begin{equation}
    v_i=L_\theta\big(P_i x_t,\,q_i,\,\mathcal{S}_i G_t,\,t\big),
    \label{eq:local}
\end{equation}
where $\mathcal{S}_i$ samples the shared features bilinearly at the tile's location, and the $v_i$ of all tiles, placed side by side, form $v_\theta$. The decoder is a shallow, attention-free U-Net, and the sampled features modulate its activations through spatially varying adaptive group normalization (AdaGN)~\cite{dhariwal2021diffusion,park2019semantic}. We train with the local velocity loss $\mathbb{E}\,\|v_i-P_iv^\star\|_2^2$ at the tile size $p$ used at inference.

The same decoder also predicts the low-resolution velocity $D_gv^\star$ of the whole field of view, which the coarse steps (\cref{sec:sampler}) use. For this prediction, the decoder takes the view $D_gx_t$, its coordinates $q_g$, and the whole feature map $\mathcal{S}_gG_t$. Since a view has the size of a tile, the coarse prediction adds almost no parameters. For a quarter of the tiles in each batch, the decoder makes this coarse prediction.

\begin{table*}[t]
\centering
\small
\caption{Reconstruction quality on the evaluation subsets (\cref{sec:baselines}): PSNR (dB) and SSIM, mean $\pm$ standard deviation over the videos. Time: seconds per frame of the reconstruction algorithm, mean $\pm$ standard deviation over the four settings; with $M$ samples, one per GPU, the latency ($M$ times this in GPU time). Best in \textbf{bold}, second \underline{underlined}.}
\label{tab:main}
\setlength{\tabcolsep}{2.4pt}
\resizebox{0.95\linewidth}{!}{%
\begin{tabular}{l cc cc cc cc c}
\toprule
& \multicolumn{2}{c}{\makecell{OCMR\\radial $R{=}16$}} & \multicolumn{2}{c}{\makecell{Brain\\1D VD $R{=}8$}} & \multicolumn{2}{c}{\makecell{Abdominal\\PROPELLER $R{=}8$}} & \multicolumn{2}{c}{\makecell{Lung CT\\$R{=}12$}} & \makecell{Time $\downarrow$} \\
\cmidrule(lr){2-3}\cmidrule(lr){4-5}\cmidrule(lr){6-7}\cmidrule(lr){8-9}
Method & PSNR$\uparrow$ & SSIM$\uparrow$ & PSNR$\uparrow$ & SSIM$\uparrow$ & PSNR$\uparrow$ & SSIM$\uparrow$ & PSNR$\uparrow$ & SSIM$\uparrow$ & (s/frame) \\
\midrule
L1-wavelet SENSE~\cite{lustig2007sparse} / FBP & 35.28\std{3.03} & .909\std{.030} & 29.09\std{2.44} & .837\std{.068} & 28.43\std{0.99} & .852\std{.027} & 25.26\std{0.84} & .314\std{.027} & \textbf{0.48}\std{0.44} \\
Spatio-temporal ADMM-TV~\cite{boyd2011admm} & 39.28\std{3.19} & .957\std{.014} & 30.35\std{2.04} & .893\std{.038} & 32.31\std{1.31} & .899\std{.020} & 36.84\std{1.06} & .961\std{.008} & 2.79\std{1.72} \\
\midrule
PaDIS~\cite{hu2024padis,padismri2025} & 40.05\std{2.75} & .954\std{.015} & 33.56\std{2.22} & .896\std{.043} & 32.62\std{1.39} & .916\std{.019} & 37.35\std{0.85} & .919\std{.009} & 219\std{150} \\
LPGC~\cite{yang2025ctpatch} & \underline{41.09}\std{3.16} & \underline{.967}\std{.012} & 35.12\std{1.53} & .941\std{.024} & 33.71\std{1.76} & .927\std{.018} & 38.47\std{0.98} & .959\std{.008} & 20.1\std{3.10} \\
Latent~\cite{varma2025medvae} + DPS~\cite{chung2023dps} & 31.42\std{1.85} & .835\std{.023} & 31.98\std{1.27} & .921\std{.023} & 29.88\std{1.19} & .861\std{.022} & 31.86\std{1.11} & .621\std{.044} & 145\std{114} \\
BC-DDS~\cite{kwon2025video} (2D prior) & 35.53\std{2.30} & .909\std{.019} & 32.99\std{1.99} & .899\std{.037} & 28.79\std{0.94} & .833\std{.016} & 37.08\std{1.06} & .948\std{.009} & 2.70\std{0.26} \\
\midrule
\textbf{\method{}} (60 steps, $M{=}1$) & 40.47\std{3.25} & .962\std{.016} & 34.88\std{1.90} & .936\std{.034} & 34.12\std{1.56} & .926\std{.018} & 39.40\std{1.12} & .972\std{.006} & \underline{0.79}\std{0.20} \\
\textbf{\method{}} (120 steps, $M{=}1$) & 40.77\std{4.27} & .966\std{.017} & \underline{35.44}\std{1.88} & \underline{.943}\std{.027} & \underline{34.66}\std{1.79} & \textbf{.931}\std{.019} & \underline{40.51}\std{1.12} & \underline{.981}\std{.005} & 2.42\std{0.39} \\
\textbf{\method{}} (120 steps, $M{=}4$) & \textbf{41.13}\std{4.40} & \textbf{.968}\std{.016} & \textbf{36.11}\std{1.91} & \textbf{.946}\std{.024} & \textbf{35.13}\std{1.85} & \underline{.929}\std{.019} & \textbf{41.10}\std{1.11} & \textbf{.984}\std{.004} & 2.47\std{0.36} \\
\bottomrule
\end{tabular}
}
\end{table*}

\subsection{Stochastic space-time tiling}
\label{sec:tessellation}
Instead of hiding seams with overlap, which multiplies the network evaluations, \sampler{} uses one non-overlapping tiling per sampling step. As in PaDIS~\cite{hu2024padis}, the image is zero-padded by $p$ pixels and covered by a grid of $p\times p$ tiles whose origin is shifted by a random offset $(a,b)\in\{0,\dots,p-1\}^2$. We extend this to time, covering the video by blocks of $W$ consecutive frames whose boundaries are shifted by a random offset $o\in\{0,\dots,W-1\}$. New offsets are drawn at every sampling step, $(a,b)_j$ per block $j$.

\subsection{Posterior sampling}
\label{sec:sampler}
\begin{algorithm}[t]
\refstepcounter{algorithm}\label{alg:retile}
\hrule height 0.8pt\vspace{2pt}
\noindent\textbf{Algorithm~\thealgorithm} \sampler{}
\vspace{2pt}\hrule\vspace{2pt}
\footnotesize
\begin{tabular}{@{}r@{\hspace{5pt}}p{0.92\linewidth}@{}}
\multicolumn{2}{@{}l}{\textbf{Require:} measurements $y$, operator $A$, noise level $\sigma$, prior $E_\theta,L_\theta$,} \\
\multicolumn{2}{@{}l}{\phantom{\textbf{Require:} }times $0=t_0<\dots<t_N=1$, switch time $t_s$} \\
{\scriptsize 1:} & $x\sim\mathcal{N}(0,I)$ \\
{\scriptsize 2:} & \textbf{for} $n=0$ \textbf{to} $N-1$ \textbf{do} \\
{\scriptsize 3:} & \quad $t\leftarrow t_n$,\; $t'\leftarrow t_{n+1}$ \\
{\scriptsize 4:} & \quad $o\sim\mathcal{U}\{0,\dots,W{-}1\}$,\; $(a,b)_j\sim\mathcal{U}\{0,\dots,p{-}1\}^2$ \hfill{\color{gray}$\triangleright$ \textit{New tiling (\cref{sec:tessellation})}} \\
{\scriptsize 5:} & \quad $G\leftarrow E_\theta(D_gx,t)$ on each block (offset $o$) \hfill{\color{gray}$\triangleright$ \textit{Eq.~\ref{eq:global}}} \\
{\scriptsize 6:} & \quad \textbf{if} $t<t_s$ \textbf{then} \hfill{\color{gray}$\triangleright$ \textit{Coarse step}} \\
{\scriptsize 7:} & \qquad $\widehat{x}_1\leftarrow D_g^{+}\big(D_gx+(1-t)\,L_\theta(D_gx,q_g,\mathcal{S}_gG,t)\big)$ \hfill{\color{gray}$\triangleright$ \textit{Eq.~\ref{eq:coarse}}} \\
{\scriptsize 8:} & \quad \textbf{else} \hfill{\color{gray}$\triangleright$ \textit{Fine step}} \\
{\scriptsize 9:} & \qquad $v\leftarrow\sum_i P_i^{\top}L_\theta(P_ix,q_i,\mathcal{S}_iG,t)$ \hfill{\color{gray}$\triangleright$ \textit{Eq.~\ref{eq:local}}} \\
{\scriptsize 10:} & \qquad $\widehat{x}_1\leftarrow x+(1-t)\,v$ \hfill{\color{gray}$\triangleright$ \textit{Eq.~\ref{eq:destination}}} \\
{\scriptsize 11:} & \quad $x^+\leftarrow\mathrm{CG}_K\big(I+\tfrac{\nu_t^2}{\sigma^2}A^HA,\ \widehat{x}_1+\tfrac{\nu_t^2}{\sigma^2}A^Hy;\ \widehat{x}_1\big)$ \hfill{\color{gray}$\triangleright$ \textit{Eq.~\ref{eq:normal}}} \\
{\scriptsize 12:} & \quad $x\leftarrow t'x^+ +(1-t')\,\xi$,\; $\xi\sim\mathcal{N}(0,I)$ \hfill{\color{gray}$\triangleright$ \textit{Eq.~\ref{eq:refresh}}} \\
{\scriptsize 13:} & \textbf{end for} \\
{\scriptsize 14:} & \textbf{return} $x$ \\
\end{tabular}
\vspace{1pt}\hrule height 0.8pt
\end{algorithm}
Each step of \sampler{} (\cref{alg:retile,fig:architecture}) has the destination--correction--re-noising form of Flower~\cite{pourya2025flower}. At time $t=t_n$, the tiles of this step's tiling (\cref{sec:tessellation}) give $v_\theta$ and the flow destination:
\begin{equation}
    \widehat{x}_1=x_t+(1-t)\,v_\theta(x_t,t).
    \label{eq:destination}
\end{equation}
The destination is refined against the measurements with the isotropic Gaussian surrogate $x_1\mid x_t\sim\mathcal{N}(\widehat{x}_1,\nu_t^2 I)$, with $\nu_t^2=(1-t)^2/\big(t^2+(1-t)^2\big)$, the posterior variance for standard Gaussian data:
\begin{equation}
    \Big(I+\tfrac{\nu_t^2}{\sigma^2}A^HA\Big)x^+=\widehat{x}_1+\tfrac{\nu_t^2}{\sigma^2}A^Hy.
    \label{eq:normal}
\end{equation}
The weight $\nu_t^2/\sigma^2$ trades the uncertainty of the destination, large early and small late, against the measurement noise. We solve \cref{eq:normal} with $K$ conjugate-gradient iterations initialized at $\widehat{x}_1$~\cite{chung2024dds} (derivation in the supplementary material). The refined destination is then re-noised to the next grid time $t'=t_{n+1}$ with fresh noise $\xi$:
\begin{equation}
    x_{t'}=t'x^+ +(1-t')\,\xi,\qquad \xi\sim\mathcal{N}(0,I).
    \label{eq:refresh}
\end{equation}
The exact solution of \cref{eq:normal} is affine in $v_\theta$. One tiling per sampling step is thus an unbiased estimate of the update with the velocity averaged over all $p^2W$ offsets~\cite{coifman1995translation,figueiredo2003em}, much as a minibatch gradient estimates the full gradient. Since each step draws a new tiling and a fresh $\xi$, tile-boundary errors in $x^+$ empirically average out over the sampling steps instead of accumulating (\cref{fig:tiling}).

\paragraph{Coarse-to-fine.} Since low frequencies are resolved first during sampling~\cite{falck2025fourier}, early on the destination is dominated by frequencies that a $p\times p$ view already resolves. As in progressive-resolution generators~\cite{jing2022subspace,jin2025pyramidal,xiao2026spectral}, for $t<t_s$ \sampler{} skips the tiles and runs the decoder once on the view,
\begin{equation}
    \widehat{x}_1=D_g^{+}\big(D_gx_t+(1-t)\,L_\theta(D_gx_t,q_g,\mathcal{S}_gG_t,t)\big),
    \label{eq:coarse}
\end{equation}
where $D_g^{+}$ zero-pads the $p\times p$ spectrum back to full resolution. For MRI, $\widehat{x}_1$ is zeroed outside the coil support, where data consistency cannot correct what a low-frequency estimate places there. The sample $x_t$, data consistency, and re-noising stay at full resolution; the tiles add detail after $t_s$. 

\begin{figure*}[t]
    \centering
    \includegraphics[width=0.90\textwidth]{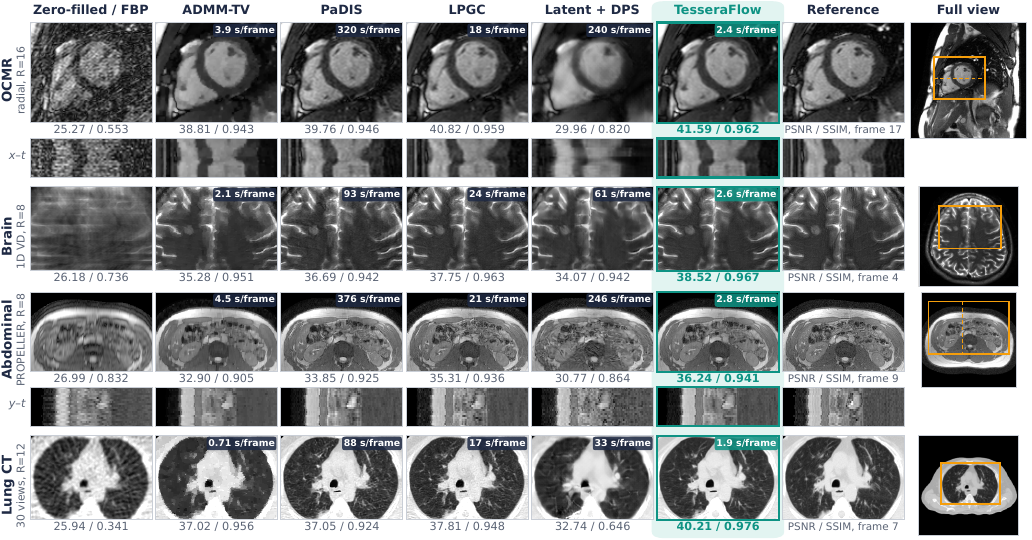}
    \caption{\textbf{Qualitative comparison}, one frame per dataset, zoomed in on the moving anatomy (\method{}: 120 steps, one sample). Last column: the reference with the zoomed region (orange) and the line (dashed) of the space--time strip. Numbers: frame PSNR (dB) / SSIM.}
    \label{fig:qualitative}
\end{figure*}

\paragraph{Tuning and multiple samples.}
We keep the annealed weight $\nu_t^2/\sigma^2$ of \cref{eq:normal}; $\sigma$, the assumed noise level, is the only weight tuned per dataset, and $t_s$ is chosen by a fixed rule (supplementary material). Running the sampler $M$ times from independent noise gives $M$ posterior samples, which share only the measurements; on $M$ GPUs they take the time of one, and their mean is the reconstruction.

\section{Experiments}
\label{sec:experiments}

We evaluate on four datasets with cardiac, head, abdominal, and respiratory motion, in two imaging modalities (MRI and CT). All four are fully sampled, and we simulate every undersampling pattern. The supplementary material details the data, baselines, and implementation, the full test sets, the paired $t$-tests, and every result below that is not in a table or figure.

\subsection{Datasets and undersampling}
\label{sec:data}

\paragraph{Cardiac cine MRI.}
OCMR~\cite{chen2020ocmr} provides fully sampled multi-coil cine k-space. We train on 34 scans, validate on one, and test on 8 held-out scans (39 slices, each a video of 10--28 frames). Each frame keeps golden-angle radial spokes at $R=16$, interleaved between frames.

\paragraph{Brain MRI with simulated head motion.}
fastMRI~\cite{zbontar2018fastmri} provides static multi-coil slices, so we simulate rigid in-plane head motion (rotation and translation) over $T=10$ frames. Each frame has its own 1D variable-density (VD) Cartesian mask at $R=8$. We train on 256 subjects, validate on 32, and test on 32, three slices each.\reviewfill

\paragraph{Abdominal MRI.}
We use fully sampled dynamic abdominal series with real physiological motion (single-slice axial scans at 3\,T, mostly 31--32 coils, one frame every 0.6--1.3\,s, $256\times256$), acquired from volunteers with written informed consent under a protocol approved by an institutional review board. We train on 8 exams, validate on 2, and evaluate on the 20 moving series of 7 held-out exams. Each frame keeps one simulated PROPELLER blade~\cite{pipe1999propeller} of 32 lines ($R=8$), resampled from the frame's fully sampled Cartesian coil k-space by a nonuniform Fourier transform and rotated by the golden angle from frame to frame, so the measurements keep the real motion and noise of the series. We evaluate the first 20 frames of every series.\reviewfill

\paragraph{Lung CT.}
The 4D-Lung collection~\cite{hugo2017lung4d,tcia4dlung,clark2013tcia} has respiratory 4D CT of 20 patients, from which we form 2D+t axial videos of 10 phases at $256\times256$. We simulate 30 parallel-beam views per phase ($R=12$), interleaved between phases, with Poisson noise. We train on 14 patients, validate on one, and test on 5, stratified by diaphragm excursion.\reviewfill

\subsection{Baselines and implementation}
\label{sec:baselines}
\begin{figure}[t]
    \centering
    \includegraphics[width=0.95\linewidth]{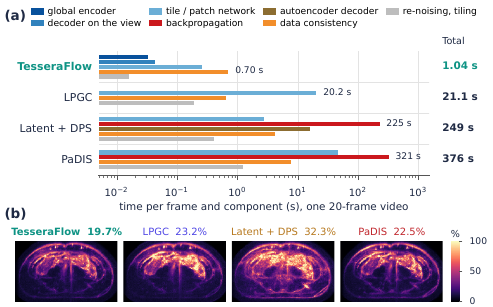}
    \caption{\textbf{Where the time goes, and temporal consistency} (abdominal MRI). (a)~Seconds per frame and component on one 20-frame video (log scale), with the total. (b)~Error of the frame-to-frame change in one video, averaged over time, in \% of the image's RMS; numbers: temporal NRMSE over the 7 videos.}
    \label{fig:efficiency}
\end{figure}

\paragraph{Classical reconstruction.}
For MRI, we run L1-wavelet SENSE per frame~\cite{lustig2007sparse} implemented in BART~\cite{uecker2015bart} with ESPIRiT sensitivities~\cite{uecker2014espirit}, and for CT, unregularized filtered backprojection in TIGRE~\cite{biguri2016tigre}. For both, spatio-temporal ADMM-TV~\cite{boyd2011admm}, set up as in~\cite{kwon2025video}, is run to convergence.

\paragraph{Generative priors.}
All generative priors are U-Nets built from the blocks of EDM~\cite{karras2022edm} and share the training data; the patch priors, including \method{}, train for 1{,}500\,kimg (thousands of training samples), the latent prior for 750\,kimg, and the BC-DDS image prior for as many optimizer steps and pixels as LPGC.
\begin{itemize}
    \item \textbf{PaDIS}~\cite{hu2024padis,padismri2025}: position-aware patch prior without global context, sampled with its published sampler.
    \item \textbf{LPGC}~\cite{yang2025ctpatch} (local patches meet global context): a patch prior with a downsampled global view concatenated to every patch input, which we train as a flow model and sample with its published conjugate-gradient sampler.
    \item \textbf{Latent + DPS}: a video diffusion prior in the latent space of MedVAE~\cite{varma2025medvae}, sampled with posterior guidance through the decoder~\cite{chung2023dps}. A public complex-valued MRI autoencoder~\cite{microsoft2024mriae} covers only brain and knee, so we encode the real and imaginary channels of each frame separately with MedVAE's 2D autoencoder and concatenate the two latents. 
    \item \textbf{BC-DDS}~\cite{kwon2025video}: the batch-consistent DDS of Kwon and Ye, a 2D image prior applied to every frame with noise shared by all frames, which we train as a flow model and sample with its published conjugate-gradient sampler.
\end{itemize}

\paragraph{Implementation.}
\method{} uses temporal blocks of $W=5$ frames and tiles and views of $p=64$ pixels, and is trained at this single tile size on one NVIDIA RTX PRO 6000 Blackwell Max-Q GPU. Reconstructions use 60 sampling steps and a single sample ($M=1$) unless stated otherwise. All times are measured on the same GPU.

\paragraph{Metrics.}
We report PSNR and SSIM on magnitude images against the fully sampled reference, and the temporal NRMSE $\|\Delta\hat x-\Delta x\|/\|x\|$ on the complex frames, where $\Delta$ takes the difference of consecutive frames and each norm is a root mean square over frames. For MRI, every reconstruction is set to zero outside the coil support, where the measurements carry no information, for all methods alike. \cref{tab:main} compares all methods on a subset of each test set.

\subsection{Reconstruction quality}
\label{sec:main}
\cref{fig:qualitative} shows one frame of each subset of \cref{tab:main}. At its default 60 steps, \method{} is the most accurate method on lung CT and has the highest PSNR on abdominal MRI. On OCMR and brain MRI, LPGC is more accurate but 21--37$\times$ slower. At the same 60 network evaluations, it is less accurate than \method{} on all four datasets. With 120 steps and the mean of four samples, \method{} matches or exceeds every generative baseline on all four datasets with about half of LPGC's GPU time, and the gain is significant everywhere except OCMR. It also has the lowest complex NRMSE, which scores the phase, on brain and abdominal MRI.
The default \method{} is more accurate than PaDIS, the latent prior, and ADMM-TV on every dataset, and faster than the converged ADMM-TV. BC-DDS is the fastest generative baseline, but its 2D prior does not model time, and it is less accurate than \method{} on every video. 

\subsection{Efficiency}
\label{sec:efficiency}
\begin{table}[t]
\centering
\footnotesize
\caption{Cost and temporal consistency on the brain MRI subset. NFE: sequential network evaluations. FPS: frames per second (with $M$ samples on $M$ GPUs, the latency). Mem.: peak memory per GPU. tNRMSE: temporal NRMSE (\cref{sec:baselines}).}
\label{tab:efficiency}
\setlength{\tabcolsep}{1.3pt}
\begin{tabular}{l r r r r r}
\toprule
Method & \makecell{NFE\\$\downarrow$} & \makecell{FPS\\$\uparrow$} & \makecell{Mem.\\(GB) $\downarrow$} & \makecell{PSNR\\$\uparrow$} & \makecell{tNRMSE\\(\%) $\downarrow$} \\
\midrule
Spatio-temporal ADMM-TV & -- & 0.48 & \textbf{0.3} & 30.35 & 21.9 \\
\midrule
PaDIS & 1040 & $<$0.02 & 33.0 & 33.56 & 16.7 \\
LPGC & 400 & $<$0.05 & 2.1 & 35.12 & 13.9 \\
Latent + DPS & 750 & $<$0.02 & 23.0 & 31.98 & 18.4 \\
BC-DDS & 100 & 0.41 & 3.4 & 32.99 & 15.5 \\
\midrule
\textbf{\method{}} (60 steps, $M{=}1$) & \textbf{60} & \textbf{1.52} & 1.6 & 34.88 & 14.0 \\
\textbf{\method{}} (120 steps, $M{=}1$) & 120 & 0.38 & 1.6 & 35.44 & 13.1 \\
\textbf{\method{}} (120 steps, $M{=}4$) & 120$\times$4 & 0.37 & 1.6 & \textbf{36.11} & \textbf{12.1} \\
\bottomrule
\end{tabular}
\end{table}
On brain MRI (\cref{tab:efficiency}), \method{} is 37--142$\times$ faster than the patch and latent priors. More steps and more samples trade time and GPUs for accuracy and temporal consistency; with both, \method{} is the most accurate method and has the lowest temporal NRMSE. On an abdominal video (\cref{fig:efficiency}a), LPGC spends most of its time on its 400 network evaluations, PaDIS and the latent prior on backpropagation. \method{} avoids both, and data consistency takes most of its time. With 60 steps, it also has the lowest temporal NRMSE on every abdominal video (\cref{fig:efficiency}b). With five-frame blocks, its cost per frame stays near 1\,s from 20 to 200 frames on one abdominal series.

\begin{figure}[t]
    \centering
    \includegraphics[width=0.95\linewidth]{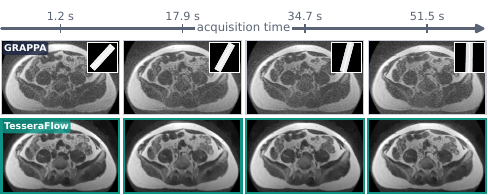}
    \caption{\textbf{\Invivo{} abdominal PROPELLER scan} at $512\times512$: frames across one 92-frame video, each from its own blade (inset). Top: each blade alone with parallel imaging (GRAPPA), with streaks and noise. Bottom: \method{}, without them.}
    \label{fig:invivo}
\end{figure}

\subsection{\texorpdfstring{\Invivo{}}{In vivo} abdominal scan}
\cref{fig:invivo} applies the abdominal model, trained at $256\times256$, to raw data from a PROPELLER scan outside our dataset, at its native $512\times512$ grid, acquired with written informed consent under an IRB-approved protocol. Each blade's skipped lines are filled from its acquired lines (GRAPPA) before data consistency. The scanner merges its 92 blades (53\,s) into one image. \method{} instead reconstructs a 92-frame video, one frame per 0.58\,s blade. Along the line of \cref{fig:teaser}c, the slow intensity changes of \method{} correlate with those of the GRAPPA and zero-filled blades (0.95 and 0.78) and agree between four samples with independent noise (0.97; supplementary material\videoonly{ and video}).

\subsection{Ablations}
\label{sec:ablations}
The supplementary material has further ablations on temporal modeling (applied frame by frame, the same architecture is less accurate on every video), the selection of $t_s$, and extra updates per sampling step.
\begin{table}[t]
\centering
\footnotesize
\caption{\textbf{Prior and sampler}: frames per second (FPS, $\uparrow$) and PSNR (dB, $\uparrow$) on each evaluation subset, from the LPGC prior~\cite{yang2025ctpatch} with its own sampler (400 evaluations), changing one design choice at a time to ours. $t_s{=}0$: no coarse steps. Highest FPS in \textbf{bold}.}
\label{tab:prior-sampler}
\setlength{\tabcolsep}{1.2pt}
\resizebox{0.95\linewidth}{!}{%
\begin{tabular}{l cc cc cc cc}
\toprule
& \multicolumn{2}{c}{OCMR} & \multicolumn{2}{c}{Brain} & \multicolumn{2}{c}{Abd.} & \multicolumn{2}{c}{Lung} \\
\cmidrule(lr){2-3}\cmidrule(lr){4-5}\cmidrule(lr){6-7}\cmidrule(lr){8-9}
Prior, sampler & FPS & PSNR & FPS & PSNR & FPS & PSNR & FPS & PSNR \\
\midrule
LPGC, original & $<$0.06 & 41.09 & $<$0.05 & 35.12 & $<$0.05 & 33.71 & $<$0.06 & 38.47 \\
LPGC, \sampler{} & 0.31 & 40.11 & 0.36 & 34.89 & 0.31 & 34.33 & 0.36 & 39.22 \\
\method{}, $t_s{=}0$ & 0.58 & 40.09 & 0.57 & 34.95 & 0.49 & 33.83 & 0.76 & 39.37 \\
\textbf{\method{}} & \textbf{1.15} & 40.47 & \textbf{1.52} & 34.88 & \textbf{0.96} & 34.12 & \textbf{1.65} & 39.40 \\
\bottomrule
\end{tabular}}
\end{table}

\paragraph{Prior and sampler.}
\cref{tab:prior-sampler} separates the two sources of speed. \sampler{} alone makes the LPGC prior~\cite{yang2025ctpatch} 6--9$\times$ faster, changing PSNR by less than 1\,dB. Under \sampler{}, our prior adds a further 3--5$\times$ at similar accuracy, about half from the coarse steps and the rest from its network, which computes the global features once per sampling step and has a shallow decoder. With 120 steps and four samples, our prior is the more accurate of the two on every dataset.

\begin{figure}[t]
    \centering
    \includegraphics[width=0.85\linewidth]{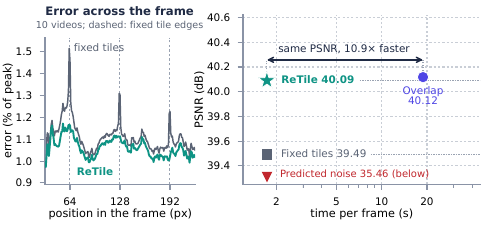}
    \caption{\textbf{Tiling.} Left: error across the frame, averaged over the 10 OCMR evaluation videos and along rows and columns. Fixed tiles err more, exactly at their edges (dashed). Right: PSNR on the OCMR evaluation subset against time per frame.}
    \label{fig:tiling}
\end{figure}
\paragraph{Tiling.}
\cref{fig:tiling} compares three ways to tile the video at each sampling step, on cardiac MRI without the coarse steps. Overlapping consensus averages overlapping tiles, fixed non-overlapping tiles keep their seams in place, and \sampler{} moves non-overlapping tiles to a new random offset each time. \sampler{} matches overlapping consensus at a tenth of the time, fixed tiles are less accurate on every video, and re-noising with the predicted instead of fresh noise costs 4.63\,dB (5.80\,dB with overlapping consensus). The random offset makes overlap unnecessary, and fresh noise is needed with any tiling.

\begin{figure}[t]
    \centering
    \includegraphics[width=0.80\linewidth]{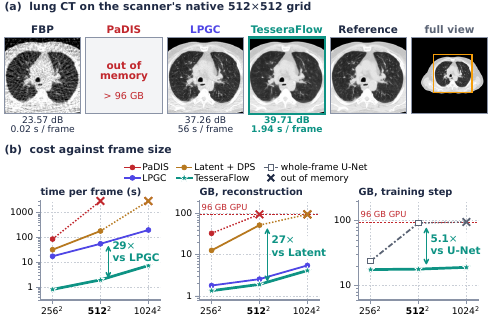}
    \caption{\textbf{Native resolution.} (a)~A lung CT test frame on the scanner's $512\times512$ grid; LPGC and \method{} are trained at $512\times512$ with their $256\times256$ recipes. (b)~Reconstruction time and peak memory, and memory of one training step.}
    \label{fig:native}
\end{figure}

\paragraph{Tiles and whole frames.} On lung CT at the native $512\times512$ grid, \method{} is more accurate than LPGC on every test video in 29$\times$ less time, while PaDIS's published sampler exceeds the GPU's memory (\cref{fig:native}a). \method{}'s training memory barely grows with the frame size, whereas a whole-frame U-Net nearly fills a 96\,GB GPU (\cref{fig:native}b). Under the same sampler, a whole-frame video U-Net is less accurate than \method{} on every lung CT and brain MRI video, 3.1--3.7$\times$ slower, and needs 3--16$\times$ the memory. \method{} trained on whole frames is at most 13\% faster but needs 2--5$\times$ the memory and is no more accurate.

\paragraph{Global encoder.} Trained identically without the encoder, the model is less accurate inside the exact coil support on every OCMR and brain MRI video, by 0.23\,dB on OCMR and 0.34\,dB on brain MRI, growing to 0.81\,dB at $R=16$.

\section{Discussion and Conclusion}
In this work, we addressed the cost of generative video priors for dynamic MRI and CT. In \method{}, a global encoder reads a low-resolution view of every frame and a single shallow decoder operates on this view during early steps and on space-time tiles during later steps, so no network processes a whole frame at full resolution. Our sampler, \sampler{}, shifts the tile grid randomly and re-noises with fresh noise at every sampling step, and we showed that one tiling per sampling step then suffices. Together, they are an order of magnitude faster than patch and latent video priors and can match or exceed the most accurate of them. From an \invivo{} abdominal scan, our framework reconstructs one $512\times512$ frame per blade, every 0.58\,s, where conventional reconstruction gives one image. Its low cost in time and memory highlights its potential for practical deployment. We leave multi-scale sampling, prospective, phase-contrast, and volumetric reconstruction to future work.

\section*{Acknowledgments}
The authors acknowledge funding from NIH grants R01EB036127 and R01HL173035.
{
    \small
    \bibliographystyle{ieeenat_fullname}
    \bibliography{main}
}

\let\arxivsetcounter\setcounter
\renewcommand{\setcounter}[2]{\ifnum\pdfstrcmp{#1}{page}=0 \else\arxivsetcounter{#1}{#2}\fi}
\clearpage
\setcounter{page}{1}
\maketitlesupplementary
\setcounter{section}{0}\renewcommand{\thesection}{\Alph{section}}\renewcommand{\theHsection}{S\Alph{section}}
\setcounter{figure}{0}\renewcommand{\thefigure}{S\arabic{figure}}\renewcommand{\theHfigure}{S\arabic{figure}}
\setcounter{table}{0}\renewcommand{\thetable}{S\arabic{table}}\renewcommand{\theHtable}{S\arabic{table}}
\setcounter{equation}{0}\renewcommand{\theequation}{S\arabic{equation}}\renewcommand{\theHequation}{S\arabic{equation}}
\setcounter{tocdepth}{1}
{\makeatletter\renewcommand*\l@section{\@dottedtocline{1}{0em}{1.6em}}\def\numberline#1{\hb@xt@\@tempdima{\hfil#1\hskip 1em}}\makeatother
\tableofcontents}
\addtocontents{toc}{\protect\setcounter{tocdepth}{1}}%
\vspace{0.6em}

This supplement gives implementation details (\cref{sec:suppl-arch}), the formulation of \sampler{} (\cref{sec:suppl-retile}), datasets and splits (\cref{sec:suppl-data}), forward operators and motion simulation (\cref{sec:suppl-ops}), baseline configurations (\cref{sec:suppl-baselines}), the data-consistency weight (\cref{sec:suppl-sigma}), the coarse-phase selection (\cref{sec:suppl-c2f}), unconditional generation (\cref{sec:suppl-uncond}), and additional results (\cref{sec:suppl-results}).

\section{Implementation details}
\label{sec:suppl-arch}
\method{} uses the same network on every dataset, with two input channels (real and imaginary part) for MRI and one for CT.

\paragraph{Global encoder.} It takes the $64\times64$ k-space crop of each full noisy frame, scaled by $64/256$, and has three levels of 96, 192, and 192 channels at 64, 32, and 16 pixels, with two residual blocks per level, single-head self-attention at $16\times16$, and a zero-initialized temporal convolution (kernel 3) after every block. Its output is a $192\times16\times16$ feature map per frame. The view comes from the same noisy state as the tiles and receives no independent noise.

\paragraph{Local decoder.} A U-Net on a $64\times64\times5$ tile with two coordinate channels, widths 96, 192, and 192, no attention, and temporal convolutions after every level. After each level of its expanding path, the global features $F$, sampled bilinearly at the tile's coordinates per frame and at that level's resolution, give $[s,b,g]=\mathrm{Conv}_{1\times1}(F)$, zero-initialized so that the update of the activations $h$ starts as the identity (GN: group normalization):
\begin{equation}
    h\leftarrow h+g\odot\mathrm{Conv}_{3\times3}\big(\mathrm{SiLU}\big((1+s)\odot\mathrm{GN}(h)+b\big)\big).
    \label{eq:suppl-modulation}
\end{equation}

\paragraph{Coarse task.} The decoder reads the $64\times64$ view with the coordinates of its pixels (the mean of the tile coordinates over the pixels that each view pixel covers) and the global features sampled there, through the same modulation, and adds a learned 384-dimensional embedding $e_c$ to its time embedding. Its target is the k-space crop of the velocity, $D_gv^\star$. Coordinates are pixel positions in the field of view scaled to $[-1,1]$. Zero-padding pixels lie outside this range, where the sampled features taper to zero. The view's coordinates span the whole field of view at four times the tile spacing, so the view reads the entire feature map and a tile reads its crop.

\paragraph{Parameters.} Encoder 5.42\,M, decoder 11.78\,M, and 384 for $e_c$, 17.20\,M in total for MRI. PaDIS and LPGC use the same 64-channel U-Net backbone with 17.39\,M parameters, \method-2D the network of \method{} without $e_c$, and the latent prior 31.71\,M (excluding MedVAE).

\paragraph{Training.} Rectified flow with $t\sim\mathcal{U}(0,1)$ and the unweighted squared error of the velocity $x-\epsilon$ (\cref{eq:path}). Adam with $\beta=(0.9,0.999)$ and weight decay $10^{-6}$, learning rate $2\times10^{-4}$ with a 40-kimg linear warm-up, gradient clipping at 1, and an EMA with a half-life of 150\,kimg, whose weights are used for sampling. Each optimizer step uses 128 five-frame $64\times64$ tiles, one random tile per clip on the zero-padded $384\times384$ canvas, and training runs for 1{,}500\,kimg (thousands of tiles) in a single stage, on one GPU with two accumulation rounds. Training takes 11.7, 10.5, 10.4, and 8.8 GPU-hours for OCMR, brain, abdominal MRI, and lung CT. Peak memory of one training forward pass (lung CT, one GPU) is 18.0, 35.8, and 71.1\,GB for 16, 32, and 64 tiles, each from a different video, and 12.0, 23.0, and 44.9\,GB for the same network on 1, 2, and 4 whole $256\times256$ clips. With the same schedule (kimg settings divided by 16, for the same optimizer steps and pixels), \method{} on whole lung CT clips reaches 39.32\,dB against 39.40\,dB on tiles (\cref{tab:prior-sampler}), and the LPGC U-Net without global and coordinate inputs 39.17\,dB with \sampler{}. In each round, the first quarter of the clips also supplies its $64\times64$ view for the coarse task, with the view loss scaled so that each view weighs as much as a tile. Timed alone, training an OCMR prior to 1{,}500\,kimg takes 11.8 GPU-hours for \method{}, 7.9 for PaDIS, and 13.6 for LPGC, and the latent prior takes 29.0 to 750\,kimg. \cref{tab:suppl-compute} adds the network operations of each prior per frame.
\begin{table}[t]
\centering
\footnotesize
\caption{Model size and compute on brain MRI ($256\times256$): parameters, network GFLOPs per evaluation and frame (matrix products, convolutions, and attention, without data consistency; for \method{} the mean over its coarse and fine steps), sequential evaluations of the published sampler (NFE, \cref{tab:efficiency}), network TFLOPs per frame, and GPU-hours to train the OCMR prior (the latent prior to 750\,kimg). $^*$Including backpropagation through the prior (and the decoder).}
\label{tab:suppl-compute}
\setlength{\tabcolsep}{3pt}
\begin{tabular}{l rrrrr}
\toprule
Prior & \makecell{Params\\(M)} & \makecell{GFLOPs\\/ eval.} & NFE & \makecell{TFLOPs\\/ frame} & \makecell{Train\\(GPU-h)} \\
\midrule
PaDIS & 17.39 & 2{,}646$^*$ & 1{,}040 & 2{,}751 & 7.9 \\
LPGC & 17.39 & 1{,}719 & 400 & 688 & 13.6 \\
Latent + DPS & 31.71 & 3{,}259$^*$ & 750 & 2{,}444 & 29.0 \\
BC-DDS & 17.4 & 854 & 100 & 85 & -- \\
\textbf{\method{}} (60 steps) & 17.20 & 190 & 60 & 11 & 11.8 \\
\bottomrule
\end{tabular}
\end{table}

\paragraph{Sampler.} 60 uniform steps on $[0,1]$ with $K=10$ conjugate-gradient iterations for \cref{eq:normal}, warm-started at $\widehat{x}_1$, and fresh noise at every step. The coarse phase always uses the step size of this grid ($\Delta t=1/60$, so 48 steps before $t_s=0.8$). With more steps, only the fine phase $[t_s,1]$ is refined, with 72 of the 120 steps. The data-consistency $\sigma$ is 0.005, 0.01, 0.005, and 0.25 for OCMR, brain, abdominal MRI, and lung CT. As in PaDIS~\cite{hu2024padis}, the tiling is a $5\times5$ grid of $64\times64$ tiles on the padded canvas, whose offset is drawn uniformly from $\{0,\dots,63\}^2$ at every network evaluation and independently for each temporal block. Temporal blocks of five frames start at an offset drawn from $\{0,\dots,4\}$. They wrap around the cardiac and lung videos. For brain and abdominal MRI, a block that would cross an end of the video is shifted inside it and keeps only its own frames, and each frame takes its velocity from one block. Overlap would multiply the decoder evaluations. Covering $T$ frames with all overlapping $W$-frame windows needs $T-W+1$ window evaluations, and spatial overlap by a stride $s$ multiplies the tile count by $(64/s)^2$.

\section{Formulation of \sampler{}}
\label{sec:suppl-retile}
\sampler{} follows the posterior-sampling construction of Flower~\cite{pourya2025flower}. We restate it in our notation and add what the tiling and the coarse phase change. Time runs from noise ($t=0$) to data ($t=1$), $x_t=(1-t)\epsilon+tx_1$ (\cref{eq:path}), on the uniform grid $0=t_0<\dots<t_N=1$.

\paragraph{Destination.} Every training pair satisfies $x_1=x_t+(1-t)(x_1-\epsilon)$, and the flow-matching optimum is $v^\star(x_t,t)=\mathbb{E}[x_1-\epsilon\mid x_t]$. The destination of \cref{eq:destination},
\begin{equation}
    \widehat{x}_1(x_t)=x_t+(1-t)\,v_\theta(x_t,t)\approx\mathbb{E}[x_1\mid x_t],
\end{equation}
is therefore an estimate of the conditional mean of the clean video given the current state (Prop.~1 of~\cite{pourya2025flower}). In a fine step, the velocity is assembled from the tiling $\omega$ drawn for this step (the temporal block offset $o$ and a spatial offset $(a,b)$ per block),
\begin{equation}
    v_\theta(x_t,t;\omega)=\sum_i P_i^{\top}L_\theta\big(P_ix_t,\,q_i,\,\mathcal{S}_iG_t,\,t\big),
    \label{eq:suppl-assembly}
\end{equation}
where the tiles $P_i$ of $\omega$ partition the zero-padded canvas and its temporal blocks partition the frames ($\sum_iP_i^{\top}P_i=I$), so each pixel receives exactly one tile velocity. In a coarse step ($t<t_s$), $\widehat{x}_1$ is given by \cref{eq:coarse} instead.

\paragraph{Local Gaussian surrogate.} The conditional $p(x_1\mid x_t)$ is intractable. As in~\cite{pourya2025flower}, \sampler{} replaces it with an isotropic Gaussian centered at the destination,
\begin{equation}
    \tilde p(x_1\mid x_t)=\mathcal{N}\big(\widehat{x}_1(x_t),\,\nu_t^2I\big),\qquad \nu_t^2=\frac{(1-t)^2}{t^2+(1-t)^2}.
\end{equation}
For standard Gaussian data, this variance is exact, since $x_1$ and $x_t$ are then jointly Gaussian with $\mathrm{Cov}(x_1,x_t)=tI$ and $\mathrm{Cov}(x_t)=\big(t^2+(1-t)^2\big)I$, so $\mathrm{Cov}(x_1\mid x_t)=\big(1-t^2/(t^2+(1-t)^2)\big)I=\nu_t^2I$. For real data it is a surrogate. It falls from 1 at $t=0$ to 0 at $t=1$, and the destination is trusted more as the state approaches the data.

\paragraph{Local MAP.} With the likelihood $p(y\mid x_1)=\mathcal{N}(Ax_1,\sigma^2I)$ of \cref{eq:measurement}, the surrogate posterior of the clean video is again Gaussian,
\begin{equation}
    \tilde p(x_1\mid x_t,y)\propto\mathcal{N}(y;Ax_1,\sigma^2I)\,\mathcal{N}(x_1;\widehat{x}_1,\nu_t^2I)=\mathcal{N}(\mu_t,\Sigma_t)
\end{equation}
(Prop.~2 of~\cite{pourya2025flower}), with
\begin{equation}
    \Sigma_t=\big(\nu_t^{-2}I+\sigma^{-2}A^HA\big)^{-1},\qquad \mu_t=\Sigma_t\big(\nu_t^{-2}\widehat{x}_1+\sigma^{-2}A^Hy\big).
\end{equation}
Its mean is also its mode, the local MAP estimate
\begin{equation}
    x^+=\mu_t=\operatorname*{arg\,min}_x\ \frac{1}{2\sigma^2}\|Ax-y\|_2^2+\frac{1}{2\nu_t^2}\|x-\widehat{x}_1\|_2^2,
    \label{eq:suppl-map}
\end{equation}
that is, the proximal step $x^+=\mathrm{prox}_{\nu_t^2F_y}(\widehat{x}_1)$ with $F_y(x)=\|Ax-y\|_2^2/(2\sigma^2)$. Multiplying its optimality condition by $\nu_t^2$ gives \cref{eq:normal}. The estimate is local in two ways. First, the prior enters only through a quadratic around the current destination, which the network re-estimates at every step. Each step therefore solves a convex least-squares problem instead of the nonconvex MAP problem of the full prior. Second, the step needs only $A$ and $A^H$. DPS differentiates $\|y-A\widehat{x}_1(x_t)\|_2^2$ with respect to $x_t$ through the network, whereas \sampler{} evaluates the network once per sampling step and never differentiates it. Since $A$ acts frame by frame, \cref{eq:suppl-map} splits into one problem per frame.

\paragraph{Truncated solve.} We run $K=10$ conjugate-gradient iterations on \cref{eq:normal}, starting at $\widehat{x}_1$~\cite{chung2024dds}. The initial residual $\beta A^H(y-A\widehat{x}_1)$, with $\beta=\nu_t^2/\sigma^2$, lies in the range of $A^H$, and $I+\beta A^HA$ maps that range into itself. All iterates therefore change $\widehat{x}_1$ only in the measured subspace and keep exactly its null-space component, which the measurements cannot inform. The exact solution has the same property, since $I+\beta A^HA$ is the identity on the null space of $A$. Within the measured subspace, stopping after $K$ iterations limits how far a step moves toward noisy measurements.

\paragraph{Re-noising.} The refined destination moves to the next grid time with fresh noise, $x_{t_{n+1}}=t_{n+1}x^++(1-t_{n+1})\xi$, $\xi\sim\mathcal{N}(0,I)$ (\cref{eq:refresh}). Under the surrogate, this update draws from the transition $\tilde p(x_{t_{n+1}}\mid x_{t_n},y)=\mathcal{N}\big(t_{n+1}\mu_t,\,t_{n+1}^2\Sigma_t+(1-t_{n+1})^2I\big)$ (Prop.~3 of~\cite{pourya2025flower}). Sampling it at every step is ancestral sampling of $p(x_t\mid y)$ when the velocity is optimal and the surrogate exact (Thm.~1 of~\cite{pourya2025flower}). Like Flower's best-performing variant ($\gamma=0$), \sampler{} drops the term $t_{n+1}^2\Sigma_t$ and re-noises the posterior mean, which gives up the exactness of Thm.~1 (it needs $\gamma=1$) for better reconstructions, as in~\cite{pourya2025flower}. At $t_N=1$ the output is the last step's $x^+$. With $M$ samples, the reconstruction is the mean of $M$ independent runs.

\paragraph{One tiling per sampling step estimates the average over all offsets.} Let $\omega$ be the tiling of a fine step ($t\geq t_s$), its temporal offset $o$ and the spatial offset $(a,b)$ of each block, all drawn uniformly, and $v_\theta(x_t,t;\omega)$ the velocity assembled from its tiles. The exact solution of \cref{eq:normal}, $x^+=\mu_t$, is affine in $\widehat{x}_1=x_t+(1-t)\,v_\theta(x_t,t;\omega)$, so for a given $x_t$
\begin{equation}
    \mathbb{E}_\omega\big[x^+(\omega)\big]=x^+(\bar v),\qquad \bar v=\mathbb{E}_\omega\big[v_\theta(x_t,t;\omega)\big],
\end{equation}
where $\bar v$ is the velocity of the tiled network averaged over all tile offsets~\cite{coifman1995translation,figueiredo2003em}. One tiling is therefore an unbiased one-sample estimate of the update with $\bar v$, as a minibatch gradient is of the full gradient, and its error $x^+(\omega)-x^+(\bar v)$ has zero mean. Overlapping consensus approximates $\bar v$ at every step with several shifted tilings, which multiplies the network evaluations. Fixed tiles use the same $\omega$ at every step, so their error is a seam pattern that repeats instead of averaging out. The truncated solve ($K=10$) is not exactly affine in its right-hand side, so for it the identity holds approximately.

\paragraph{Why fresh noise lets the tiles stop overlapping.} With noise predicted from the velocity, $\widehat{\epsilon}=x_t-t\,v_\theta$, and without data consistency, the update would be
\begin{equation}
    t'\widehat{x}_1+(1-t')\,\widehat{\epsilon}=x_t+(t'-t)\,v_\theta(x_t,t;\omega),
\end{equation}
an Euler step of the probability-flow ODE. The state would carry every past tile-boundary error forward deterministically, and the next velocity would be evaluated on it. With fresh noise, the next state depends on the past only through $t'x^+$, its noise part is independent of every tiling, and the next step draws an independent $\omega'$. Because the offsets are uniform over a full tile period, every pixel lies next to a tile edge with the same probability, $1-(62/64)^2$ per sampling step, so no position accumulates boundary errors. For the cine and lung videos, whose temporal blocks wrap around, the block offsets act in the same way on the frames. On OCMR, inside the exact coil support (\cref{tab:suppl-tiling}), random tiles reach 40.16\,dB in 40\,s against 40.19\,dB in 431\,s for overlapping consensus, fixed tiles reach 39.81\,dB, and re-noising with the predicted instead of fresh noise 35.24\,dB, or 34.07\,dB with overlapping consensus. Predicted noise thus fails without fixed seams as well, so fresh noise matters for the sampler as a whole, not only against seams.

\section{Datasets and splits}
\label{sec:suppl-data}
\begin{table}[t]
\centering
\footnotesize
\caption{Evaluation settings. Each video is one slice followed over time, and training and test data come from disjoint scans, exams, and patients. $T$ is the number of frames per video; parentheses give the evaluation subset of Table~\ref{tab:main}. Sampling: golden-angle radial at $R=16$ (OCMR), 1D VD masks at $R=8$ (brain), one 32-line PROPELLER blade per frame ($R=8$, abdominal), and 30 views per phase ($R=12$, lung CT).}
\label{tab:data}
\setlength{\tabcolsep}{2pt}
\begin{tabular}{llcccc}
\toprule
Dataset & Trained on & $T$ & \makecell{Train\\frames} & \makecell{Test\\videos} & \makecell{Test\\frames} \\
\midrule
OCMR cine MRI & OCMR & 10--28 & 2665 & 39 (10) & 866 (229) \\
Brain, head motion & fastMRI brain T2 & 10 & 7680 & 96 (32) & 960 (320) \\
Abdominal MRI & Abdominal & 20 & 2465 & 20 (7) & 400 (140) \\
Lung CT & 4D-Lung & 10 & 13250 & 40 (5) & 400 (50) \\
\bottomrule
\end{tabular}
\end{table}

\cref{tab:data} summarizes the four evaluation settings.
\paragraph{OCMR.} All slices of a scan belong to the same split. We hold out 20\% of the scans in each stratum of field strength (1.5\,T, 3\,T) and acquisition type (single slice, multi-slice stack) with a fixed seed. This gives 34 training scans (117 slices) and 9 held-out scans. One of them, fs\_0096 (3\,T, one slice of 27 frames), is used for validation, which leaves 8 evaluation scans (39 slices, 10--28 frames). Its field of view is aliased (OCMR metadata), which the single-map coil model cannot represent, and \method{} and the classical baselines reach only 22--24\,dB on it. Because of this aliasing, all OCMR hyperparameters of every method, including $t_s$, are selected on three training slices instead (\cref{sec:suppl-baselines}). The evaluation scans are fs\_0009, fs\_0027, fs\_0032, fs\_0045, fs\_0057, fs\_0074, fs\_0090, and fs\_0091. The scans have 18--38 coil channels (18--34 in the evaluation scans), used without coil compression.
\paragraph{Brain MRI.} The brain data are the multi-coil T2-weighted scans of fastMRI~\cite{zbontar2018fastmri}, split by scan into 256 training, 32 validation, and 32 test subjects, with slices 3, 6, and 9 of each scan. No scan contributes to more than one split. The scans have 2--20 coil channels (4--20 in the test subjects), used without coil compression.

\paragraph{Abdominal MRI.} Series have 30--300 frames per slice, and all series of an exam belong to the same split. Training uses 8 exams (28 series, 2465 frames), validation 7 series from 2 exams, and testing 20 series from 7 exams. The longest validation series, from a pediatric exam, is used for the scaling study with $T$ up to 200. Five of the test exams keep their first 25 frames.

PROPELLER sampling is applied retrospectively. Each frame keeps one blade of 32 phase-encode lines, resampled from its fully sampled coil k-space with a nonuniform Fourier transform, and the blade angle advances by the golden angle from frame to frame, so $T$ frames complete about $T/13$ revolutions. Taken from the acquired k-space, the measurements keep the acquisition noise and its deviation from the coil model. The series are fully sampled single-slice axial balanced SSFP acquisitions at 3\,T (TR/TE 3.9--5.9/1.7--2.2\,ms, flip angle 45--60$^\circ$), during free breathing or breath-holds, with one frame every 0.6--1.3\,s. The exams have 31--32 coil channels, except one training exam with 8, used without coil compression. In preprocessing, the k-space of each series (acquired matrices from $200\times140$ to $320\times224$) is center-cropped or zero-padded to $256\times256$, and the field of view is then enlarged by 1.2 (the image is zero-padded to $307\times307$ and its k-space cropped back to $256\times256$, so each pixel covers 1.2 times its original width and the body lies inside the circle that every blade angle covers). Every series is normalized by one scalar (the 99th percentile of a low-resolution root-sum-of-squares image from the central 24 lines), and coil maps are estimated for every frame with ESPIRiT (calibration radius 24, one map).
\paragraph{Lung CT.} The videos come from the 4D-Lung collection. Patients are sorted by diaphragm excursion (7.8--44.5\,mm) into six strata, and one patient per stratum is drawn with a fixed seed and held out, which leaves 14 training patients. One held-out patient (14.5\,mm excursion) is used for validation and the other five for evaluation. Since the evaluation set includes the patient with the largest excursion, we report every metric per patient. The scans are reconstructed on a $512\times512$ grid with 0.98--1.16\,mm pixels and 3\,mm slices. We convert them to HU, set the padding outside the scan field to air, clip to $[-1024,3071]$\,HU, remove the couch, keep a slab of 96 slices centred on the lungs, and average $2\times2$ pixel blocks, which gives $256\times256$ frames with 1.95--2.32\,mm pixels. Images are linear attenuation relative to water, $x=1+\mathrm{HU}/1000$, clipped to $[0,3]$, so the forward model is linear without an offset. Each evaluation patient contributes eight test videos, the axial slices at evenly spaced positions in the central 80\% of its 96 slices (40 videos and 400 frames). The same videos are used for every method.
\paragraph{Evaluation subsets.} The generative baselines need up to 1{,}040 network evaluations per five-frame chunk, so all methods in \cref{tab:main} are compared on a fixed positional subset of each test set: every fourth OCMR test slice (10 videos), the middle slice of each brain test subject (32), the middle series of each abdominal test exam (7), and the middle evaluated slice of each lung CT patient (5).

\begin{figure*}[t]
    \centering
    \includegraphics[width=\textwidth]{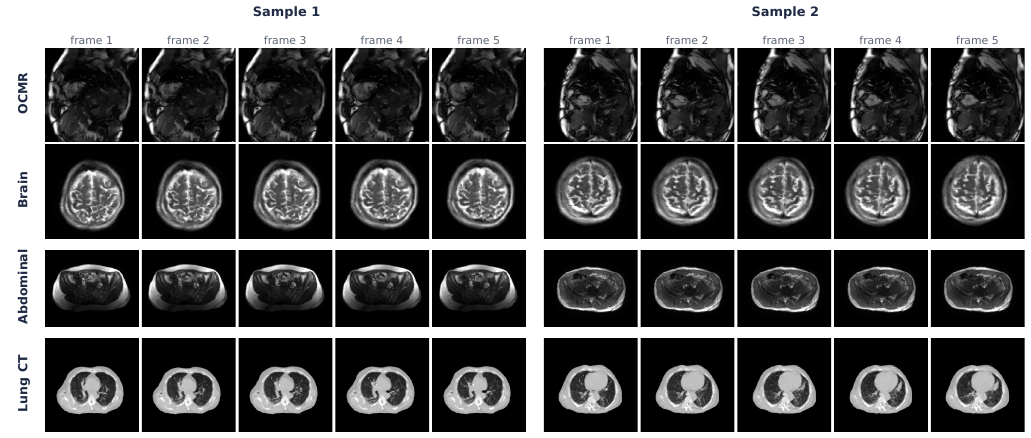}
    \caption{Unconditional samples of \method{} (coarse phase, $t_s=0.5$, 100 Heun steps, one seed, no selection): the first two 5-frame clips drawn on each dataset, all frames. Magnitude (MRI) or attenuation (CT), each clip scaled to its own range.}
    \label{fig:suppl-uncond}
\end{figure*}
\begin{figure*}[t]
    \centering
    \includegraphics[width=\textwidth]{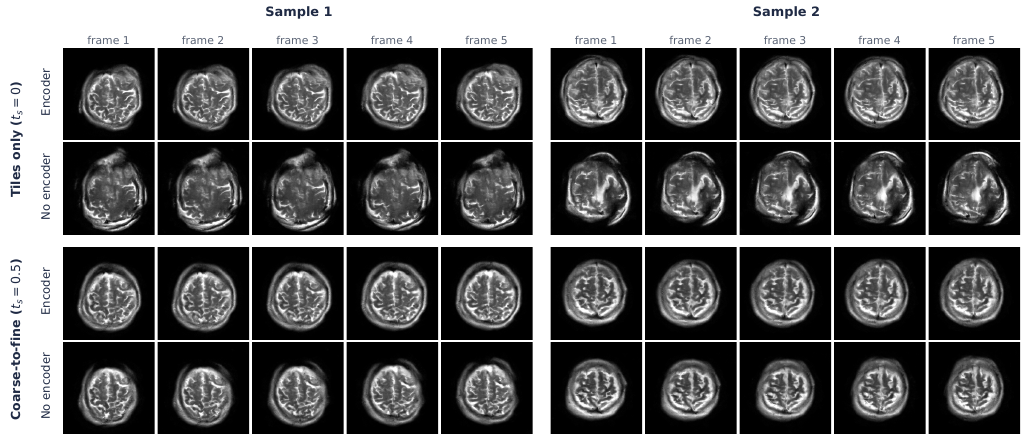}
    \caption{Unconditional brain samples with and without the global encoder (both at 1{,}500\,kimg; 100 Heun steps, same noise), from the tiles alone ($t_s=0$) and with the coarse phase ($t_s=0.5$): the first two 5-frame clips of each run, all frames. Magnitude, each clip scaled to its own range.}
    \label{fig:suppl-encoder}
\end{figure*}
\section{Forward operators and motion simulation}
\label{sec:suppl-ops}
Every method reconstructs with the operator that simulates its measurements (nonuniform FFT or CT projector), which favors all methods alike, but the MRI measurements come from the acquired coil k-space and keep its noise and its deviation from the coil model. All methods share the same coil maps, which ESPIRiT estimates from each fully sampled frame (calibration region 24), not from the acquired calibration lines alone. OCMR: frame $t$ takes spokes $16t,\dots,16t+15$ of a golden-angle sequence that starts at a different index for every video, so that consecutive frames interleave. For radial and PROPELLER data, $A^HA$ in \cref{eq:normal} uses a Toeplitz embedding, with two FFTs on a twice-oversampled grid per conjugate-gradient iteration. LPGC uses the same embedding in its conjugate-gradient steps. PaDIS and the latent prior never apply $A^HA$. Their guidance differentiates the residual norm of the denoised estimate (for the latent prior, of its decoding) through the prior by automatic differentiation, which needs $A$ itself. On the abdominal video of \cref{fig:efficiency}, their forward-operator calls take 2\% of the time and backpropagation 85--90\%, so the embedding would barely change their times. Brain: all slices of a subject follow one rigid trajectory over ten frames, with rotation $4^\circ\sin(2\pi t/8+\phi_r)$ and a translation of $6\sin(2\pi t/8+\phi_s)$ pixels of the $320\times320$ crop along a direction $\beta$, where $\phi_r$, $\phi_s$, and $\beta$ are drawn per subject from a fixed seed. The motion is applied to the multi-coil k-space, so the coil sensitivities move with the head, and ESPIRiT maps are estimated for every frame. In a real scan, the coils stay in place while the head moves. This gives mean steps of $2^\circ$ and 3 pixels per frame. Each frame is rotated in k-space by a nonuniform adjoint transform on rotated coordinates and translated by a linear phase ramp, and its coil sensitivities are re-estimated with ESPIRiT~\cite{uecker2014espirit}. Each frame is sampled with a 1D variable-density mask at $R=8$ with 8 calibration lines, redrawn for every frame. Abdominal MRI: the motion is that of the acquired series. Each blade covers 32 lines of the 256-sample readout (a band of relative width 0.125) at the golden angle $n\pi/\phi \bmod \pi$ for frame $n$, and the nonuniform FFT uses Kaiser-Bessel interpolation with twofold oversampling. CT: each phase is projected at $(k+(f \bmod 10)/10)\,\pi/30$, $k=0,\dots,29$, so the ten phases interleave into 300 distinct angles, by rotating the image (bilinear interpolation) and summing over rows into 256 detector bins. The adjoint is the exact transpose, computed by automatic differentiation. Measurements are Poisson counts with $I_0=10^5$ photons per bin, log-transformed back to line integrals.

\section{Baseline configurations}
\label{sec:suppl-baselines}
The classical baselines are selected on validation data: for brain MRI, one video from each of three validation subjects; for OCMR, the three training slices used for our $\sigma$, because its single validation scan has an aliased field of view, which the single-map coil model cannot represent, and selects much stronger regularization ($\lambda=0.03$ for L1-wavelet and $0.009$ for ADMM-TV, at the edge of the grid). Every grid is extended once when its optimum lies at an edge, and the selection rule is that of $\sigma$ (\cref{sec:suppl-sigma}).
\paragraph{L1-wavelet SENSE and FBP.} BART~\cite{uecker2015bart} runs \texttt{pics -i 100 -g -S -R W:3:0:$\lambda$} (FISTA with Daubechies-2 wavelets and random cycle spinning~\cite{figueiredo2003em}, with the per-frame trajectory and \texttt{-e} added for non-Cartesian data) on non-overlapping 5-frame blocks, with the per-frame coil maps that every method uses (ESPIRiT~\cite{uecker2014espirit}, calibration radius 24, one map). BART rescales the data internally, so its $\lambda$ is not comparable with that of ADMM-TV. The grid $\lambda\in\{1,3\}\times10^{\{-4,\dots,-2\}}\cup\{10^{-1}\}$ selects $10^{-3}$ on OCMR and brain MRI and $3\cdot10^{-4}$ on abdominal MRI. CT filtered backprojection reconstructs each phase from its own 30 views with TIGRE~\cite{biguri2016tigre}. Of the Ram-Lak, Shepp-Logan, cosine, Hamming, and Hann filters, validation selects Hann.

\paragraph{Spatio-temporal ADMM-TV.} It minimizes $\tfrac12\|\mathcal{A}X-Y\|_2^2+\lambda\|DX\|_1$ with anisotropic finite differences $D=[D_t,D_h,D_w]$ along time and both image axes and the exact measurement operator $\mathcal{A}$, solved by ADMM~\cite{boyd2011admm} from zero initialization (300 iterations with 10 warm-started inner conjugate-gradient iterations). The differences are circular, complex differences are thresholded on their magnitude, and $\rho=0.1$ for MRI and 1 for CT. $\mathcal{A}$ is the masked orthonormal FFT for brain MRI, a nonuniform FFT with a Toeplitz normal operator for radial and PROPELLER data, and the projector for CT. The selected $\lambda$ is $3\cdot10^{-4}$, $10^{-3}$, $3\cdot10^{-4}$, and 0.3 for OCMR, brain, abdominal MRI, and lung CT. On training or validation items at the selected $\lambda$, 1{,}000 instead of 300 iterations change the mean PSNR by $-0.020$, $+0.001$, $-0.003$, and $+0.001$\,dB on OCMR, brain, abdominal MRI, and lung CT (at most 0.09\,dB on any item).
\paragraph{Generative baselines.} All generative baselines keep their published samplers. Only their data-consistency strength is set per dataset, as follows. PaDIS~\cite{hu2024padis,padismri2025} uses 104 (MRI) or 100 (CT) noise levels with 10 Langevin steps each, $\alpha=0.5\sigma^2$, and its published step $\zeta$ converted exactly to our image and operator units (3.50, 7.93, 2.44, and 0.100 for OCMR, brain, abdominal MRI, and lung CT). On brain and abdominal MRI, it follows PaDIS-MRI~\cite{padismri2025}. Each Langevin step evaluates the denoiser at $x+\sigma\epsilon$ with fresh complex Gaussian noise $\epsilon$ of variance $1/2$ per real and imaginary part and adds Langevin noise of the same variance, and training uses Adam with learning rate $10^{-4}$, the 10{,}000-kimg linear warm-up of its code, and an EMA half-life of 500\,kimg, on patches of 16, 32, and 64 pixels. On cardiac MRI, PaDIS trained with PaDIS-MRI's recipe (on patches of 32, 64, and 96 pixels, with 32 pixels of zero padding) generates plausible unconditional samples (\cref{fig:suppl-uncond-compare}) but diverged on parts of three of the ten evaluation videos. There, as on lung CT, PaDIS is trained with the learning rate of the PaDIS paper~\cite{hu2024padis}, $2\times10^{-4}$, and the warm-up and EMA of our priors. A sweep of $\zeta\times\{0.5,1,2\}$ on one chunk per dataset improved PSNR by at most 0.27\,dB over the published $\zeta$, which we kept. On lung CT, tuning on three validation chunks selected $\zeta\times0.5$. LPGC~\cite{yang2025ctpatch} uses 200 DDIM steps with $\eta=0.8$ and two recurrent noising rounds. We train the LPGC prior as a flow model, and the same prior serves both samplers of \cref{tab:prior-sampler}. Its DDIM sampler runs exactly in this parameterization, because the flow path $x_t=tx_1+(1-t)\epsilon$ divided by $s_t=\sqrt{t^2+(1-t)^2}$ is a variance-preserving diffusion path with signal scale $t/s_t$ and noise scale $(1-t)/s_t$, and the clean and noise estimates come from the predicted velocity. LPGC models static 3D volumes, and each of its recurrent rounds partitions the volume into non-overlapping patches at a random offset along all three axes. For videos, time takes the place of the third axis. Each round covers the video with non-overlapping five-frame blocks from a random temporal offset, wrapped around the periodic cardiac and lung CT videos and shifted inside the others, as in \method{}. Averaging the prior over every overlapping five-frame window instead changes PSNR by $-0.17$ to $+0.02$\,dB per dataset at 2.1--3.9$\times$ the time. With \sampler{}, the LPGC prior is within 0.50\,dB of \method{} without its coarse phase and 1.6--2.1$\times$ slower (\cref{tab:prior-sampler}). The number $K$ of conjugate-gradient iterations, which the paper does not state, is 2 for OCMR and 1 for brain MRI, and 5 for abdominal MRI and lung CT, where a validation sweep found it within 0.2\,dB of the best. Latent + DPS~\cite{varma2025medvae,chung2023dps} uses 750 steps ($\sigma$ from 80 to 0.002) on five-frame chunks, with the guidance step selected per dataset (3, 3, 10, and 0.03). Tuning on OCMR and abdominal MRI uses the first five-frame chunk of each video. The autoencoder is MedVAE's 2D model with $4\times$ downsampling and 3 latent channels. Complex MRI frames are scaled by $s$ (0.65, 0.65, 1.0, and 0.8 for OCMR, brain, abdominal MRI, and lung CT) and their real and imaginary parts encoded separately, giving 6 latent channels, normalized per channel with training statistics. The prior is a video diffusion model on the $64\times64$ latents of five-frame clips.

\paragraph{BC-DDS.} Kwon and Ye~\cite{kwon2025video} do not name their method; BC-DDS abbreviates the batch-consistent DDS of their Sec.~3.2. Their pretrained image prior is replaced by the whole-frame U-Net of the lung CT control (17.4\,M parameters, no global encoder, no tiles, no coordinates), trained as a flow model on single $256\times256$ frames of the same training data, with the optimizer steps and pixels per step of LPGC. At each of 100 steps, every frame passes through the prior on its own, all frames in one batch; $l$ conjugate-gradient iterations on $A^HAx=A^Hy$, started at the clean estimate, run over the whole video; and the result is re-noised with a share $\sqrt{1-\eta^2}$ of the predicted noise and $\eta=0.8$ of one noise frame shared by all frames, redrawn at every step. The same shared noise starts the video. These are their settings for spatio-temporal degradations. Since the flow path divided by $s_t$ is a variance-preserving path, this is the update of their code exactly. We follow the code where it differs from the paper, in the split of the noise and in a last conjugate-gradient pass after the final step. The steps are uniform in flow time instead of in the index of their noise schedule. Their method has no data-consistency weight, so $l$ sets the data-consistency strength. We select it from $\{1,2,3,5,10\}$, the values of the DDS ablation~\cite{chung2024dds} and of LPGC's selection, on the selection data of the other baselines, which gives 3, 2, 5, and 5 for OCMR, brain, abdominal MRI, and lung CT.

\paragraph{\method-2D.} The same network without the coarse task is trained on single frames for 7{,}500\,kimg with 640 single-frame tiles per step, so that it trains on as many frames and takes as many optimizer steps as \method{} (128 five-frame tiles per step, 1{,}500\,kimg); EMA half-life and warm-up are scaled by five. It is sampled frame by frame with the settings of \method{}, without the coarse phase.

\section{Data-consistency weight}
\label{sec:suppl-sigma}
\begin{figure}[t]
    \centering
    \includegraphics[width=\linewidth]{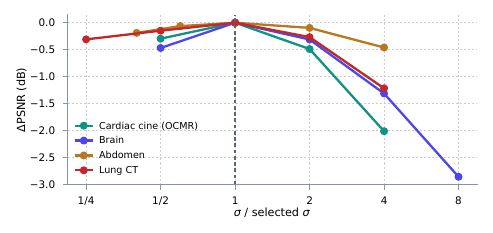}
    \caption{PSNR of the final \method{} ($t_s=0.8$, 60 steps) on the selection data (training or validation scans, never test data) against the data-consistency $\sigma$, relative to the selected value. The selected $\sigma$ is the best on every dataset, and halving or doubling it costs at most 0.49\,dB.}
    \label{fig:suppl-sigma}
\end{figure}
The data-consistency $\sigma$ of \method{} was selected per dataset and the rule used for the classical baselines (highest mean PSNR, SSIM breaking ties within 0.05\,dB), and the grid is extended whenever the optimum lies at its edge. The OCMR hyperparameters of every method were selected with full-field-of-view metrics, before the coil-support mask of the evaluation was extended to OCMR. Selection and evaluation otherwise follow the same protocol. OCMR: 60 steps, $\sigma\in\{0.001,0.002,0.003,0.005,0.01,0.02,0.03,0.05,0.1\}$, giving $\sigma=0.005$.
Brain MRI: 60 steps, $\sigma\in\{0.005,0.01,0.02,0.04,0.08\}$, giving $\sigma=0.01$. Abdominal MRI: 60 steps, $\sigma\in\{0.002,0.003,0.005,0.01,0.02\}$, giving $\sigma=0.005$. Lung CT: 60 steps, $\sigma\in\{0.0625,0.125,0.25,0.5,1\}$, giving $\sigma=0.25$. \method-2D uses the same $\sigma$ as \method{}. \cref{fig:suppl-sigma} repeats every sweep with the final model, on the same items and grids.

\begin{table}[t]
\centering
\footnotesize
\caption{Reconstruction time per frame (s, $\downarrow$) on the evaluation subsets: sampler or solver time per video divided by its number of frames, on one NVIDIA RTX PRO 6000 Blackwell Max-Q. Table~\ref{tab:main} reports the mean. Best in \textbf{bold}.}
\label{tab:time-per-frame}
\setlength{\tabcolsep}{3pt}
\begin{tabular}{l ccccc}
\toprule
Method & OCMR & Brain & Abd. & Lung & Mean \\
\midrule
L1-wavelet SENSE / FBP & \textbf{0.71} & \textbf{0.22} & \textbf{0.97} & \textbf{0.01} & \textbf{0.48} \\
Spatio-temporal ADMM-TV & 3.86 & 2.08 & 4.49 & 0.71 & 2.79 \\
\midrule
PaDIS & 320 & 93.1 & 376 & 88.0 & 219 \\
LPGC & 18.1 & 24.1 & 21.0 & 17.3 & 20.1 \\
Latent + DPS & 240 & 61.4 & 246 & 33.3 & 145 \\
BC-DDS & 2.80 & 2.44 & 3.02 & 2.54 & 2.70 \\
\method-2D & 1.93 & 1.56 & 2.03 & 1.56 & 1.77 \\
\midrule
\textbf{\method{}} (60 steps, $M{=}1$) & 0.87 & 0.66 & 1.05 & 0.61 & 0.79 \\
\textbf{\method{}} (120 steps, $M{=}4$) & 2.44 & 2.67 & 2.80 & 1.98 & 2.47 \\
\bottomrule
\end{tabular}
\end{table}

\begin{figure*}[t]
    \centering
    \includegraphics[width=\textwidth]{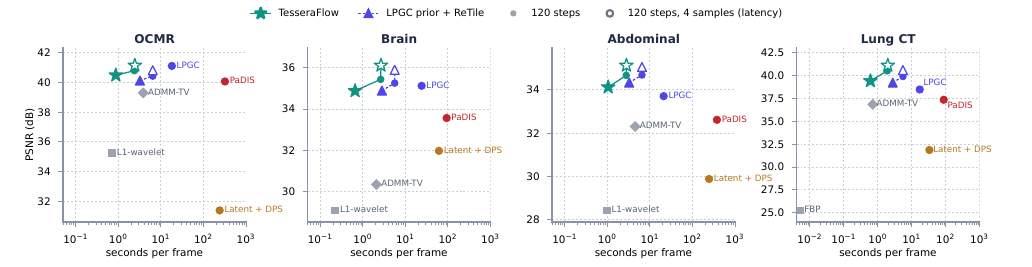}
    \caption{Quality against reconstruction time per frame on the four evaluation subsets (data of \cref{tab:main} and \cref{tab:time-per-frame}). Stars: \method{}; triangles: the LPGC prior under \sampler{}. Filled: 60 steps; dot: 120 steps; open: 120 steps and four samples on four GPUs, at the latency of one. The 120-step LPGC runs shared their GPUs, so their time is twice that of its timed 60-step run, since all of its steps run on patches.}
    \label{fig:suppl-quality-time}
\end{figure*}
\begin{figure*}[t]
    \centering
    \includegraphics[width=\textwidth]{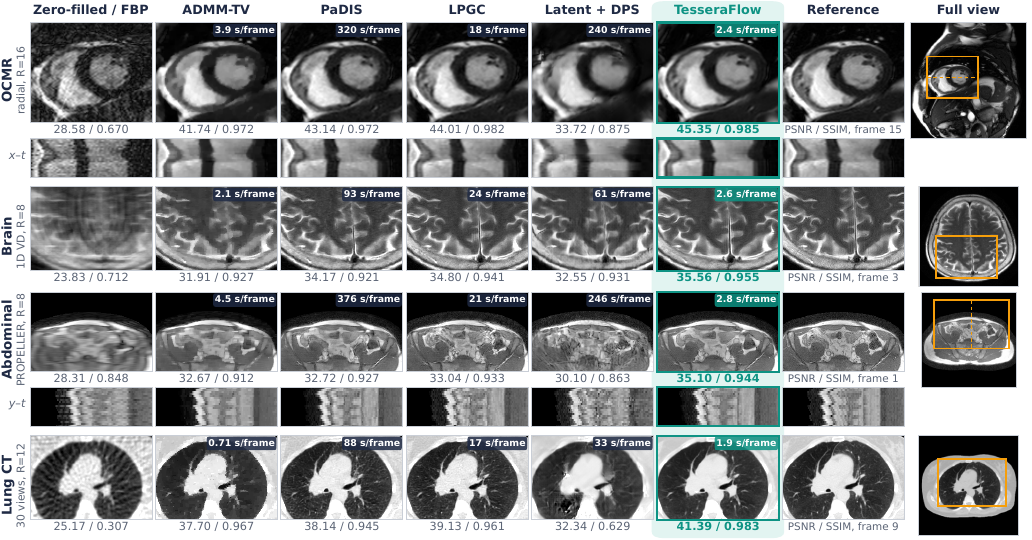}
    \caption{Qualitative comparison on other test videos (layout of \cref{fig:qualitative}; \method{} with 120 steps and one sample). PSNR (dB) / SSIM of the frame, best in bold; badges: reconstruction time per frame.}
    \label{fig:suppl-qualitative}
\end{figure*}
\begin{figure*}[t]
    \centering
    \includegraphics[width=\textwidth]{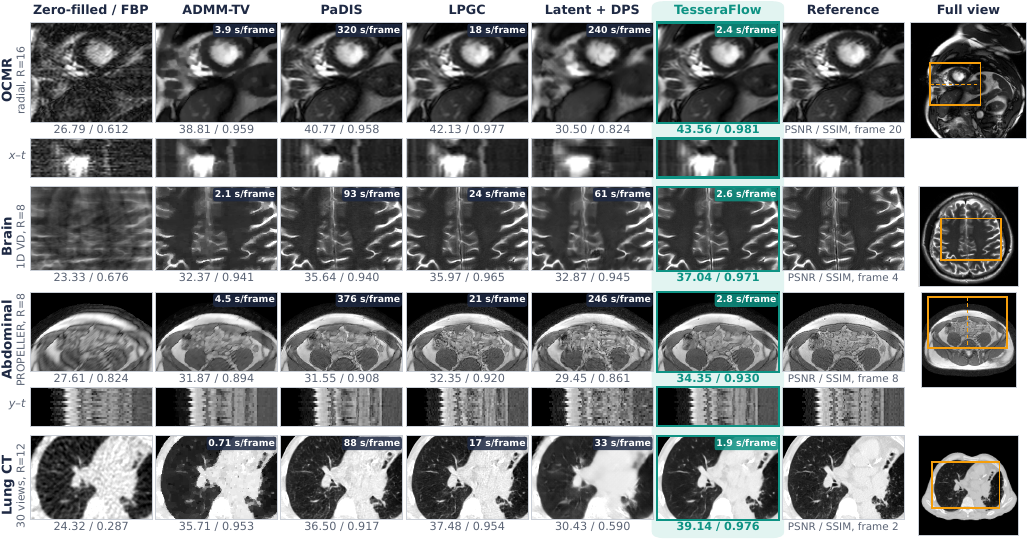}
    \caption{Qualitative comparison on further test videos (layout of \cref{fig:qualitative}; \method{} with 120 steps and one sample). PSNR (dB) / SSIM of the frame, best in bold; badges: reconstruction time per frame.}
    \label{fig:suppl-qualitative2}
\end{figure*}
\begin{table*}[t]
\centering
\footnotesize
\caption{\method{}, LPGC, BC-DDS, and the classical baselines on the full test sets (OCMR 39, brain 96, abdominal 20, and lung CT 40 videos); PaDIS and Latent + DPS only on the evaluation subsets of \cref{tab:main}. PSNR in dB / SSIM. Best in \textbf{bold}.}
\label{tab:fulltest}
\setlength{\tabcolsep}{4pt}
\begin{tabular}{l cc cc cc cc}
\toprule
& \multicolumn{2}{c}{OCMR} & \multicolumn{2}{c}{Brain} & \multicolumn{2}{c}{Abdominal} & \multicolumn{2}{c}{Lung CT} \\
\cmidrule(lr){2-3}\cmidrule(lr){4-5}\cmidrule(lr){6-7}\cmidrule(lr){8-9}
Method & PSNR$\uparrow$ & SSIM$\uparrow$ & PSNR$\uparrow$ & SSIM$\uparrow$ & PSNR$\uparrow$ & SSIM$\uparrow$ & PSNR$\uparrow$ & SSIM$\uparrow$ \\
\midrule
L1-wavelet SENSE / FBP & 35.42 & .915 & 28.83 & .837 & 28.33 & .846 & 25.73 & .317 \\
Spatio-temporal ADMM-TV & 39.46 & .960 & 30.38 & .897 & 32.07 & .893 & 36.48 & .947 \\
\midrule
LPGC~\cite{yang2025ctpatch} & 41.31 & .970 & 34.98 & .940 & 33.66 & .925 & 37.59 & .940 \\
BC-DDS~\cite{kwon2025video} (2D prior) & 35.46 & .914 & 32.93 & .901 & 28.70 & .828 & 36.73 & .931 \\
\midrule
\textbf{\method{}} (60 steps, $M{=}1$) & 40.73 & .966 & 34.68 & .936 & 33.84 & .923 & 39.33 & .962 \\
\textbf{\method{}} (120 steps, $M{=}1$) & 41.19 & .971 & 35.23 & .942 & 34.41 & \textbf{.928} & 40.03 & .972 \\
\textbf{\method{}} (120 steps, $M{=}4$) & \textbf{41.56} & \textbf{.972} & \textbf{35.89} & \textbf{.946} & \textbf{34.86} & .926 & \textbf{40.61} & \textbf{.975} \\
\bottomrule
\end{tabular}
\end{table*}

\begin{table}[t]
\centering
\scriptsize
\caption{Paired $t$-tests of PSNR on the evaluation subsets of \cref{tab:main}: mean difference \method{} $-$ baseline (dB) and two-sided $p$-value, with the videos of each subject averaged first (4 OCMR scans, 32 brain subjects, 7 abdominal exams, 5 lung CT patients). Bold: $p<0.05$; no correction for multiple comparisons.}
\label{tab:suppl-ttest}
\setlength{\tabcolsep}{2pt}
\begin{tabular}{l cccc}
\toprule
Baseline & OCMR & Brain & Abd. & Lung \\
\midrule
\multicolumn{5}{l}{\method{} (60 steps, $M{=}1$) $-$} \\
L1-wavelet / FBP & \textbf{+5.45} (.002) & \textbf{+5.79} ($<$.001) & \textbf{+5.69} ($<$.001) & \textbf{+14.14} ($<$.001) \\
ADMM-TV & \textbf{+1.25} (.018) & \textbf{+4.53} ($<$.001) & \textbf{+1.81} ($<$.001) & \textbf{+2.56} ($<$.001) \\
PaDIS & +0.45 (.27) & \textbf{+1.32} ($<$.001) & \textbf{+1.51} ($<$.001) & \textbf{+2.05} ($<$.001) \\
LPGC & \textbf{-0.61} (.038) & \textbf{-0.24} (.016) & \textbf{+0.42} (.021) & +0.93 (.063) \\
Latent + DPS & \textbf{+9.09} ($<$.001) & \textbf{+2.91} ($<$.001) & \textbf{+4.24} ($<$.001) & \textbf{+7.54} ($<$.001) \\
BC-DDS & \textbf{+4.92} ($<$.001) & \textbf{+1.89} ($<$.001) & \textbf{+5.34} ($<$.001) & \textbf{+2.32} ($<$.001) \\
\midrule
\multicolumn{5}{l}{\method{} (120 steps, $M{=}4$) $-$} \\
L1-wavelet / FBP & \textbf{+6.29} (.006) & \textbf{+7.02} ($<$.001) & \textbf{+6.69} ($<$.001) & \textbf{+15.84} ($<$.001) \\
ADMM-TV & \textbf{+2.10} (.046) & \textbf{+5.77} ($<$.001) & \textbf{+2.82} ($<$.001) & \textbf{+4.26} ($<$.001) \\
PaDIS & \textbf{+1.30} (.043) & \textbf{+2.55} ($<$.001) & \textbf{+2.51} ($<$.001) & \textbf{+3.75} ($<$.001) \\
LPGC & +0.24 (.62) & \textbf{+0.99} ($<$.001) & \textbf{+1.42} ($<$.001) & \textbf{+2.63} ($<$.001) \\
Latent + DPS & \textbf{+9.93} ($<$.001) & \textbf{+4.14} ($<$.001) & \textbf{+5.24} ($<$.001) & \textbf{+9.24} ($<$.001) \\
BC-DDS & \textbf{+5.77} ($<$.001) & \textbf{+3.12} ($<$.001) & \textbf{+6.34} ($<$.001) & \textbf{+4.02} ($<$.001) \\
\bottomrule
\end{tabular}
\end{table}

\paragraph{Paired tests.} \cref{tab:suppl-ttest} tests the differences of \cref{tab:main} over subjects. With 120 steps and four samples, \method{} beats LPGC on 6 of 10, 29 of 32, 7 of 7, and 5 of 5 videos on OCMR, brain MRI, abdominal MRI, and lung CT, and is more accurate than every baseline on every dataset at $p<0.05$, except LPGC on OCMR ($p=0.62$), where the evaluation subset holds four scans. With its default 60 steps, it is less accurate than LPGC on OCMR and brain MRI and more accurate than every other baseline, at $p<0.05$ except PaDIS on OCMR and LPGC on lung CT.
\section{Coarse-to-fine sampling}
\label{sec:suppl-c2f}
Two rules, fixed before the models were trained, select $t_s$ before any test run. First, the coarse task must leave the tiles unharmed. At $t_s=0$, the validation mean must stay within 0.15\,dB of the same prior trained without the coarse task. Second, $t_s$ is the largest value in $\{0.7,0.8\}$ that costs at most 0.05\,dB PSNR and 0.002 SSIM against $t_s=0$. On the validation splits of brain MRI (three videos), abdominal MRI (seven series), and lung CT (eight slices of one patient), the tiles change by $+0.007$, $-0.004$, and $-0.050$\,dB, and $t_s=0.8$ changes PSNR by $-0.035$, $+0.158$, and $+0.032$\,dB. For OCMR, as for every other OCMR hyperparameter (\cref{sec:suppl-sigma}), the three OCMR training slices are used, and the tiles change by $+0.014$\,dB and $t_s=0.8$ by $+0.023$\,dB. This gives $t_s=0.8$ on every dataset. On the evaluation subsets, it changes PSNR by $-0.07$, $+0.38$, $+0.29$, and $+0.03$\,dB on brain, OCMR, abdominal, and lung CT, at 2.7, 2.0, 1.9, and 2.2$\times$ shorter sampling time.

The 120-step rows keep this coarse step size and give the other 72 steps to $[t_s,1]$ when the second rule, applied at 120 steps, admits it. It does on abdominal MRI, OCMR, and lung CT ($+0.10$, $+0.50$, and $+0.76$\,dB). On brain MRI, $t_s=0.8$ costs 0.0023 SSIM, so the rule selects $t_s=0.7$, with 43 coarse steps (the grid point at 0.7 falls just below it in single precision; $+0.20$\,dB, SSIM $-0.0011$). With the same 120 steps and four samples, the LPGC prior under \sampler{} reaches 40.80, 35.90, 35.05, and 40.59\,dB on OCMR, brain MRI, abdominal MRI, and lung CT (\cref{fig:suppl-quality-time}), 0.33, 0.22, 0.07, and 0.51\,dB below \method{}, which is more accurate on 6, 28, 4, and 5 of the 10, 32, 7, and 5 videos.

\paragraph{Why the coarse task sits in the decoder.} An earlier version predicted the coarse velocity with a separate head (0.39\,M parameters) on the global features, trained after the prior had converged with all other weights frozen. On the brain validation videos, that head passes $t_s=0.7$ ($+0.004$\,dB) but not 0.8 ($-0.21$\,dB), while the shared decoder passes 0.8 ($-0.035$\,dB, SSIM $-0.0002$). Its coarse estimate is also about ten times more accurate. The error of the low band on the unmeasured k-space, through the coil operator, is 1--5\% of the band's energy for the decoder and 10--32\% for the head. At $t_s=0.7$ the measurements and the fine steps correct even the head's estimate, so the two reach the same quality there, but without measurements they do not (\cref{sec:suppl-uncond}).

\begin{table}[t]
\centering
\footnotesize
\caption{Component ablations on the three OCMR training slices used for tuning (radial, $R=16$; first five frames; 60 steps, $t_s=0.8$). Each row changes one component of the full model, shown in the first row. $\sigma$ is the data-consistency noise level and $K$ the number of conjugate-gradient iterations.}
\label{tab:ablation-components}
\setlength{\tabcolsep}{3pt}
\begin{tabular}{l cc}
\toprule
Variant & PSNR$\uparrow$ & SSIM$\uparrow$ \\
\midrule
\textbf{\method{} (full)}: $\sigma=0.005$, $K=10$, weight $\nu_t^2$ & 41.23 & .957 \\
\midrule
Mixed tile sizes in training & 40.90 & .956 \\
\midrule
Linear data weight $1-t$, best $\sigma=0.02$ & 41.01 & .951 \\
$\sigma\times0.5$ & 40.93 & .949 \\
$\sigma\times2$ & 40.74 & .955 \\
$K=5$ CG iterations & 40.96 & .957 \\
$K=50$ CG iterations & 40.50 & .949 \\
\midrule
$M=4$ samples (mean) & \textbf{41.87} & \textbf{.961} \\
\bottomrule
\end{tabular}
\end{table}

\begin{figure}[t]
    \centering
    \includegraphics[width=\linewidth]{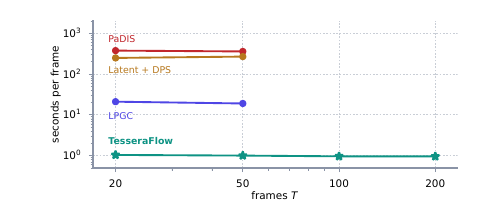}
    \caption{Seconds per frame against the number of frames $T$ on the abdominal scaling series (baselines at $T=20$ and 50). Every method works on 5-frame windows, so its cost per frame is constant in $T$.}
    \label{fig:suppl-scaling}
\end{figure}
\begin{table}[t]
\centering
\footnotesize
\caption{Tiling, global encoder, and re-noising on the OCMR and brain MRI evaluation subsets (10 and 32 videos, 60 steps, without the coarse phase). Re-noising draws fresh noise or reuses the noise predicted by the network. The model without the global encoder is trained identically, with the same data and kimg. PSNR and SSIM inside the exact coil support: the squared error and the SSIM map are averaged over the pixels where the root-sum-of-squares of the coil maps exceeds 5\% of its maximum, without the 2-pixel dilation of the scoring mask of \cref{tab:main} and \cref{fig:tiling}. Time: sampler seconds per video (OCMR).}
\label{tab:suppl-tiling}
\setlength{\tabcolsep}{2.5pt}
\begin{tabular}{c l l c c r}
\toprule
Global & Tiling & Re-noise & PSNR$\uparrow$ & SSIM$\uparrow$ & Time (s) $\downarrow$ \\
\midrule
\multicolumn{6}{@{}l}{\textit{OCMR}} \\
\checkmark & random (\sampler{}) & fresh & 40.16 & 0.958 & 40 \\
\checkmark & fixed tiles and blocks & fresh & 39.81 & 0.956 & 39 \\
\checkmark & overlapping consensus & fresh & 40.19 & 0.959 & 431 \\
\checkmark & random & predicted & 35.24 & 0.867 & 40 \\
\checkmark & overlapping consensus & predicted & 34.07 & 0.840 & -- \\
\addlinespace
-- & random & fresh & 39.93 & 0.957 & 36 \\
-- & fixed tiles and blocks & fresh & 39.67 & 0.955 & 36 \\
-- & overlapping consensus & fresh & 39.97 & 0.957 & 371 \\
\midrule
\multicolumn{6}{@{}l}{\textit{Brain MRI}} \\
\checkmark & random (\sampler{}) & fresh & 33.40 & 0.920 & -- \\
\checkmark & fixed tiles and blocks & fresh & 33.00 & 0.916 & -- \\
\checkmark & overlapping consensus & fresh & 33.33 & 0.919 & -- \\
\checkmark & random & predicted & 25.23 & 0.599 & -- \\
\addlinespace
-- & random & fresh & 33.06 & 0.917 & -- \\
-- & fixed tiles and blocks & fresh & 32.69 & 0.913 & -- \\
-- & overlapping consensus & fresh & 33.01 & 0.917 & -- \\
\bottomrule
\end{tabular}
\end{table}

\begin{figure*}[t]
    \centering
    \includegraphics[width=\textwidth]{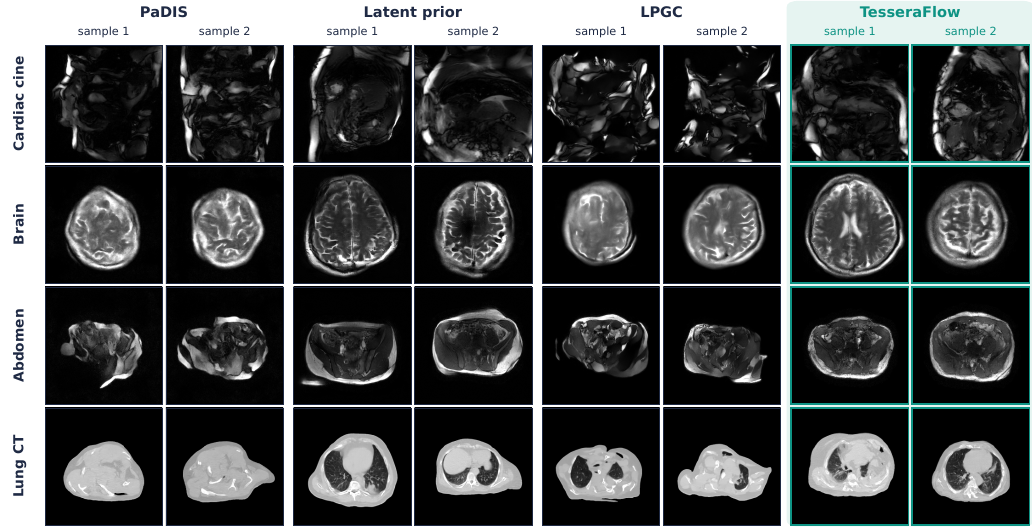}
    \caption{\textbf{Unconditional samples of each prior on the four datasets}, two each, no selection (middle frame of a 5-frame clip). PaDIS and LPGC use their own samplers, the latent prior 100 Heun steps (\method{}: coarse-to-fine with $t_s=0.5$). Magnitude (MRI) or attenuation (CT), each panel scaled to its own range.}
    \label{fig:suppl-uncond-compare}
\end{figure*}
\begin{figure*}[t]
    \centering
    \includegraphics[width=\textwidth]{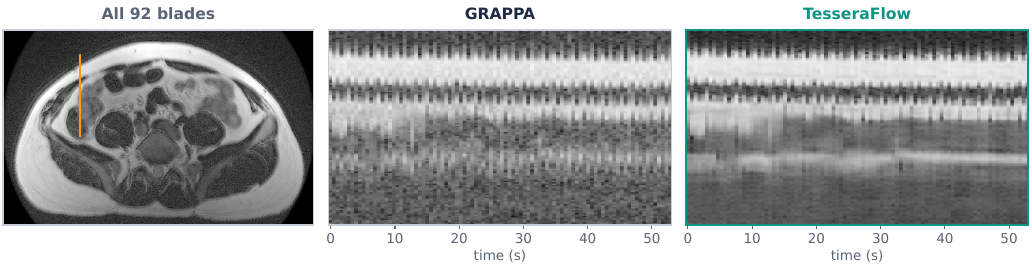}
    \caption{\textbf{x-t view of the \invivo{} scan} of \cref{fig:invivo}: a line through bowel loops on the other side of the abdomen from the line of \cref{fig:teaser}c, chosen by the same rule on the zero-filled data and marked on the conventional reconstruction, one image from all 92 blades (left), over all 92 frames, from each frame's own blade with GRAPPA (one display gain for all frames) and from \method{}.}
    \label{fig:suppl-invivo-xt}
\end{figure*}
\section{Unconditional generation}
\label{sec:suppl-uncond}
Without measurements, \method{} generates videos by integrating its prior velocity from noise with Heun's method, with a new random tiling at every evaluation, in 5-frame clips (the training window), with the coarse phase ($t_s=0.5$) and 100 steps. On brain MRI (\cref{fig:suppl-uncond-compare}), 200 instead of 100 steps change the samples by 2\% (relative $\ell_2$). The separate head of \cref{sec:suppl-c2f} fails here, because without measurements nothing corrects its low-frequency estimate, and its coarse-to-fine samples keep only the body outline. \cref{fig:suppl-uncond} shows the first two clips drawn on each dataset. Brain, abdominal, and lung anatomy is coherent and consistent across frames, while the cardiac samples reproduce the chest outline and tissue texture but no coherent heart. \cref{fig:suppl-encoder} compares samples with and without the global encoder (\cref{sec:suppl-results}).

\paragraph{PaDIS.} On lung CT, PaDIS generates with its annealed Langevin sampler (100 noise levels, 10 steps each). On MRI, we use a noise-controlled variant of it, from $\sigma=10$ with $\alpha=0.5\sigma^2$ (40 noise levels with 6 Langevin steps each on brain MRI and 3 on cardiac MRI; 50 levels with 2 steps on abdominal MRI), where each step adds the drift $\tfrac{\alpha}{2}(D(x;\sigma)-x)/\sigma^2$ and then Gaussian noise of variance $\sigma^2-s^2$, with $s$ the standard deviation of $x$ on the zero padding, so that the noise in $x$ stays at $\sigma$. \cref{fig:suppl-uncond-compare} compares two samples of each prior on all four datasets. LPGC generates with its own recurrent-noising sampler without data consistency, with the two rounds and $\eta=0.8$ of its reconstruction setting and 100 steps.

\section{Additional results}
\label{sec:suppl-results}
\paragraph{Why a global encoder.}
The local decoder receives one $64\times64\times5$ tile and its pixel coordinates, which tell it where the tile lies but not what the rest of the frame contains, such as the anatomy, the contrast, and the motion state. Without this context, the tiles can agree on these only through the shared state, from step to step. The global encoder provides this context from the low-frequency view of every frame of the temporal block, once per sampling step for all tiles (\cref{eq:global,eq:local}). To test whether the tiles need it, we trained the model identically without the encoder (10.76\,M parameters). Its decoder still performs the coarse task on the view but has no global features to read. On brain MRI, with both models at 1{,}500\,kimg, the encoder-free model loses 0.36\,dB at $t_s=0$ on three validation videos (0.33--0.37\,dB on each, SSIM $-0.004$, and 0.31\,dB at 750\,kimg). The coarse phase does not close this gap. The encoder-free model passes the second rule of \cref{sec:suppl-c2f} at $t_s=0.7$ and 0.8, like the full model, but still loses 0.36\,dB at $t_s=0.7$, because the coarse phase covers only the steps before $t_s$ and the tiles resolve the detail after it. Without measurements and at $t_s=0$, its tiles alone do not form a coherent head. The outline is distorted and bright blobs appear, while the tiles of the full model stay coherent. With the coarse phase, both models generate plausible anatomy (\cref{fig:suppl-encoder}). Inside the exact coil support (\cref{tab:suppl-tiling}), the full model is more accurate with every tiling. On brain MRI, it gains 0.34, 0.32, and 0.31\,dB with \sampler{}, overlapping consensus, and fixed tiles, on all 32 videos each, and on OCMR 0.23, 0.22, and 0.14\,dB, on 10, 10, and 9 of the 10 videos. 

\begin{figure}[t]
    \centering
    \includegraphics[width=\linewidth]{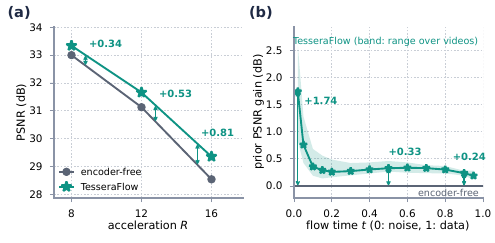}
    \caption{\textbf{Where the global encoder helps} on the 32 brain MRI evaluation videos (arrows: gain over the encoder-free model, inside the exact coil support). (a)~Reconstruction with the default sampler against the acceleration. (b)~One step of the tiles alone, without measurements: the reference is noised to $x_t$, and the plot shows the PSNR gain of the clean estimate $x_t+(1-t)v$ against the reference, as a function of the flow time; both models on identical noise and tilings; band: range over the videos.}
    \label{fig:suppl-encoder-gain}
\end{figure}

\paragraph{Sparser measurements.} \cref{fig:suppl-encoder-gain} repeats this comparison on the 32 brain MRI evaluation videos, with only the acceleration changed. Inside the exact coil support, with the default sampler ($t_s=0.8$), the encoder's gain grows from 0.34\,dB at $R=8$ to 0.53 and 0.81\,dB at $R=12$ and 16, on every video at every $R$, and it is larger at $R=16$ than at $R=8$ on all 32 videos; at $t_s=0$ it is 0.34, 0.48, and 0.67\,dB. With the default sampler, the encoder also lowers the temporal NRMSE (\cref{tab:suppl-temporal}) along with its PSNR gain, from 20.45 to 18.98\% at $R=16$, on all 32 videos. Without measurements, on identical noise and tilings, the tiles of the full model give a more accurate clean estimate than those of the encoder-free model on every video at every noise level, by 1.74\,dB at $t=0.02$ and 0.2--0.3\,dB from $t=0.15$ on.

\paragraph{Temporal modeling.} \method-2D is the same architecture trained on single frames, with as many frames and optimizer steps, and applied frame by frame with \sampler{}. Since it has no coarse prediction, we compare it with \method{} at $t_s=0$ on the evaluation subsets. It is less accurate on every video of all four datasets, by 4.12, 0.95, 4.22, and 2.18\,dB on OCMR, brain MRI, abdominal MRI, and lung CT.

\paragraph{Tiling on brain MRI.} The tiling ablation of \cref{fig:tiling} carries over to brain MRI (\cref{tab:suppl-tiling}). Inside the exact coil support, random non-overlapping tiles are 0.07\,dB more accurate than overlapping consensus, fixed tiles lose 0.40\,dB, and re-noising with the predicted noise loses 8.17\,dB, each on all 32 videos.

\begin{table}[t]
\centering
\footnotesize
\caption{Sampling steps versus updates per sampling step on the OCMR evaluation subset (10 videos, without the coarse phase). $N$ sampling steps with $K$ updates each, every update with a new random tiling, conjugate gradients, and fresh re-noising, cost $NK$ network evaluations. PSNR and SSIM with the scoring mask of \cref{tab:main}. Last column: videos on which $K=1$ at the same evaluations is more accurate. The $60\times1$ row is \sampler{} as in \cref{fig:tiling}.}
\label{tab:suppl-inner}
\setlength{\tabcolsep}{4pt}
\begin{tabular}{c c c c c c}
\toprule
Evaluations & $N$ & $K$ & PSNR$\uparrow$ & SSIM$\uparrow$ & $K=1$ better \\
\midrule
60 & 60 & 1 & \textbf{40.09} & \textbf{0.960} & -- \\
60 & 30 & 2 & 39.83 & 0.957 & 10/10 \\
60 & 15 & 4 & 39.17 & 0.947 & 10/10 \\
\addlinespace
120 & 120 & 1 & \textbf{40.48} & \textbf{0.964} & -- \\
120 & 60 & 2 & 40.40 & 0.964 & 10/10 \\
\bottomrule
\end{tabular}
\end{table}

\paragraph{Updates per sampling step.} LPGC re-noises and denoises twice per sampling step, and PaDIS takes several Langevin steps per noise level, each with a new random tiling. To test whether \sampler{} gains from such inner updates, we run $K-1$ extra updates at the same flow time before each step, each with a new random tiling, conjugate gradients, and fresh re-noising back to $t$, so that $N$ steps cost $NK$ network evaluations (\cref{tab:suppl-inner}). At equal evaluations, more steps with one tiling each are more accurate on all 10 OCMR evaluation videos, by 0.26 and 0.92\,dB against $K=2$ and $K=4$ at 60 evaluations and by 0.07\,dB against $K=2$ at 120. The inner updates help at a fixed number of steps, by 0.31\,dB from $K=1$ to $K=2$ at 60 steps, but less than the same evaluations spent on steps (0.39\,dB), consistent with one tiling per sampling step being an unbiased estimate of the update averaged over all offsets.

\paragraph{Components.}
On the OCMR training slices used for tuning (\cref{tab:ablation-components}), single-tile training beats mixed tile sizes by 0.33\,dB at equal kimg, on each of the three slices, $\nu_t^2$ beats a linear weight $1-t$ by 0.22\,dB at each one's best $\sigma$, halving or doubling $\sigma$ costs less than 0.5\,dB, and ten conjugate-gradient iterations beat a converged solve ($K=50$) by 0.73\,dB with a fifth of the iterations, as early stopping regularizes the measurement fit. Sampling the mixed-size model, whose view stays at $64\times64$, with $32\times32$ instead of $64\times64$ tiles changes PSNR by less than 0.01\,dB on the OCMR evaluation videos, at 4\% more time. A model trained with $p=32$, whose view shrinks to $32\times32$ as well, with the optimizer steps and pixels per step of $p=64$ and sampled with its settings, is 0.18\,dB less accurate on the brain evaluation subset, on 31 of 32 videos, and 0.10\,dB less accurate on the abdominal subset (4 of 7 exams, not significant). At $256\times256$, a plain whole-frame U-Net is 0.23 and 0.62\,dB less accurate than \method{} on lung CT and brain MRI, on every video, and \method{} trained on whole frames loses 0.08\,dB on lung CT (4 of 5 videos).
\cref{fig:suppl-quality-time} shows the quality--time plane of every evaluation subset, \cref{fig:suppl-qualitative,fig:suppl-qualitative2} show other test videos, \cref{tab:fulltest} compares \method{} with LPGC, BC-DDS, and the classical baselines on the full test sets, and \cref{fig:suppl-scaling} shows the time per frame against the video length.

\begin{table}[t]
\centering
\footnotesize
\caption{Equal budget on lung CT: every generative prior at 60 sequential network evaluations, the default of \method{}, on the evaluation subset (5 videos), each baseline with its data-consistency strength retuned for this budget on validation data (\cref{sec:suppl-results}). Steps $\times$ rounds: noise levels $\times$ Langevin steps (PaDIS), DDIM steps $\times$ recurrent rounds (LPGC). Time per frame as in \cref{tab:time-per-frame}. Best in \textbf{bold}.}
\label{tab:suppl-budget}
\setlength{\tabcolsep}{3pt}
\begin{tabular}{l l c c c}
\toprule
Prior & Steps $\times$ rounds, setting & PSNR$\uparrow$ & SSIM$\uparrow$ & Time (s) $\downarrow$ \\
\midrule
PaDIS & $60\times1$, $\zeta{\times}16$ & 20.62 & .142 & 5.28 \\
LPGC & $30\times2$, $K{=}10$, $\eta{=}0$ & 38.11 & .948 & 2.62 \\
Latent + DPS & 60, guidance${\times}2$ & 22.52 & .667 & 2.66 \\
\midrule
\textbf{\method{}} & 60, $t_s{=}0.8$ & \textbf{39.40} & \textbf{.972} & \textbf{0.61} \\
\bottomrule
\end{tabular}
\end{table}

\paragraph{Equal budget.} \cref{tab:suppl-budget} gives the patch and latent priors the 60 sequential network evaluations of \method{} on lung CT, instead of the 400 to 1{,}040 of its published sampler (\cref{tab:efficiency}). Each data-consistency strength is retuned for this budget on the three lung CT validation slices (protocol of \cref{sec:suppl-baselines}), over $K\in\{1,2,3,5,10\}$ and $\eta\in\{0,0.8\}$ for LPGC and over 1, 2, 4, and 8 times the published-budget $\zeta$ (0.05, with 6 levels of 10 steps or 60 of one) or guidance step (0.03) for PaDIS and the latent prior. PaDIS selects the largest step tried. PaDIS and the latent prior fail at this budget, and LPGC stays 1.30\,dB and 0.024 SSIM below \method{} at 4.3 times its time. LPGC is timed at this budget; the times of PaDIS and the latent prior scale their timed published-budget runs (\cref{tab:time-per-frame}) with the evaluations, and profiles at two step counts (\cref{fig:efficiency}) bound their fixed cost by 0.7\% of a 60-step run. On the MRI subsets, LPGC with its own data consistency and only the step count lowered to 60 evaluations ($30\times2$) is also less accurate than \method{} with 60 steps: 37.73 against 40.47\,dB on OCMR (8 of 10 videos), 34.02 against 34.88\,dB on brain MRI (24 of 32), and 33.76 against 34.12\,dB on abdominal MRI (7 of 7), at $p<0.005$ over subjects.

\begin{table}[t]
\centering
\footnotesize
\caption{Complex NRMSE (\%, $\downarrow$) on the MRI evaluation subsets: $\|x-x_\mathrm{ref}\|_2/\|x_\mathrm{ref}\|_2$ over the complex frames of a video (zero outside the coil support, as for PSNR), mean $\pm$ standard deviation over the videos. Best in \textbf{bold}, second \underline{underlined}.}
\label{tab:suppl-nrmse}
\setlength{\tabcolsep}{2pt}
\begin{tabular}{l ccc}
\toprule
Method & OCMR & Brain & Abdominal \\
\midrule
L1-wavelet SENSE~\cite{lustig2007sparse} & 15.14\std{4.12} & 20.65\std{3.95} & 29.43\std{2.77} \\
ADMM-TV~\cite{boyd2011admm} & 9.81\std{3.45} & 17.06\std{2.63} & 19.48\std{2.75} \\
PaDIS~\cite{hu2024padis,padismri2025} & 9.31\std{2.80} & 12.28\std{2.45} & 19.50\std{3.20} \\
LPGC~\cite{yang2025ctpatch} & \textbf{8.20}\std{3.28} & \underline{10.16}\std{1.52} & 18.31\std{3.19} \\
Latent~\cite{varma2025medvae} + DPS~\cite{chung2023dps} & 23.06\std{2.38} & 14.50\std{2.13} & 27.12\std{4.14} \\
BC-DDS~\cite{kwon2025video} & 14.19\std{2.96} & 12.86\std{2.20} & 29.49\std{3.99} \\
\method-2D (frame by frame) & 13.59\std{3.71} & 11.28\std{2.36} & 26.42\std{2.54} \\
\midrule
\method{} (60 steps, $M{=}1$) & 8.82\std{3.36} & 10.36\std{2.01} & \underline{16.59}\std{2.71} \\
\method{} (120 steps, $M{=}4$) & \underline{8.49}\std{4.91} & \textbf{9.02}\std{1.84} & \textbf{15.03}\std{2.93} \\
\bottomrule
\end{tabular}
\end{table}

\paragraph{Complex NRMSE.} PSNR and SSIM score magnitude images. \cref{tab:suppl-nrmse} scores the complex images and therefore also the phase. \method{} with four samples has the lowest NRMSE on brain and abdominal MRI, and LPGC on OCMR, 0.29 points below it. The temporal NRMSE (\cref{tab:suppl-temporal}) has the same best method on each MRI subset; on lung CT, \method{} is lowest.
\begin{table}[t]
\centering
\scriptsize
\caption{Temporal NRMSE (\%, $\downarrow$) of \cref{tab:efficiency} on every subset, mean over the videos (real images for lung CT). TF: \method{} (steps $\times$ samples). Best in \textbf{bold}, second \underline{underlined}.}
\label{tab:suppl-temporal}
\setlength{\tabcolsep}{1.6pt}
\begin{tabular}{l ccccccccc}
\toprule
 & \makecell{L1 /\\FBP} & \makecell{ADMM-\\TV} & PaDIS & LPGC & \makecell{Latent\\+ DPS} & \makecell{BC-\\DDS} & \makecell{TF-\\2D} & \makecell{\textbf{TF}\\60} & \makecell{\textbf{TF}\\120$\times$4} \\
\midrule
OCMR & 18.77 & 10.37 & 11.28 & \textbf{8.83} & 14.37 & 16.93 & 16.92 & 10.03 & \underline{9.14} \\
Brain & 26.61 & 21.94 & 16.75 & \underline{13.89} & 18.37 & 15.51 & 15.50 & 13.98 & \textbf{12.08} \\
Abdominal & 33.55 & 22.05 & 22.47 & 23.19 & 32.31 & 33.67 & 31.27 & \underline{19.72} & \textbf{18.46} \\
Lung CT & 20.01 & 6.45 & 5.61 & 6.26 & 6.09 & 6.34 & 6.56 & \underline{4.27} & \textbf{3.58} \\
\bottomrule
\end{tabular}
\end{table}

\paragraph{Autoencoder round trip.} A latent prior can only represent what its autoencoder reproduces. \cref{tab:suppl-autoencoder} encodes and decodes the fully sampled reference of every MRI evaluation video with MedVAE~\cite{varma2025medvae}, as in the latent baseline, and with the public complex-valued MRI autoencoder~\cite{microsoft2024mriae}, with its own preprocessing: real and imaginary channels, each frame divided by the 99.5th percentile of its magnitude, decoded from the latent mean. Rotating its input by a global phase changes PSNR by at most 0.3\,dB, and scaling the input by 0.5 to 2 improves on its own normalization by at most 0.01\,dB. It is 1.4\,dB better than MedVAE on brain MRI, the anatomy it was trained on, and 1.2 and 1.5\,dB worse on cardiac and abdominal MRI. Our brain videos come from scans of the fastMRI training split, which this autoencoder may have seen; its model card does not state its training split.
\begin{table}[t]
\centering
\footnotesize
\caption{Autoencoder round trip on the MRI evaluation subsets (number of videos in parentheses): the fully sampled reference of each video is encoded and decoded, without measurements, and scored as in \cref{tab:main} (magnitude, zero outside the coil support), mean $\pm$ standard deviation over the videos. MedVAE encodes the real and imaginary parts separately, as in the latent baseline; the complex-valued autoencoder takes them as two channels. Better in \textbf{bold}.}
\label{tab:suppl-autoencoder}
\setlength{\tabcolsep}{2pt}
\begin{tabular}{l cc cc}
\toprule
 & \multicolumn{2}{c}{MedVAE~\cite{varma2025medvae}} & \multicolumn{2}{c}{Complex-valued~\cite{microsoft2024mriae}} \\
\cmidrule(lr){2-3}\cmidrule(lr){4-5}
Dataset & PSNR$\uparrow$ & SSIM$\uparrow$ & PSNR$\uparrow$ & SSIM$\uparrow$ \\
\midrule
OCMR (10) & \textbf{38.67}\std{1.50} & \textbf{.976}\std{.006} & 37.52\std{2.24} & .958\std{.010} \\
Brain (32) & 33.14\std{1.38} & \textbf{.950}\std{.016} & \textbf{34.52}\std{1.08} & .949\std{.019} \\
Abdominal (7) & \textbf{34.37}\std{1.33} & \textbf{.945}\std{.013} & 32.90\std{1.55} & .917\std{.016} \\
\bottomrule
\end{tabular}
\end{table}

\begin{figure}[t]
    \centering
    \includegraphics[width=\linewidth]{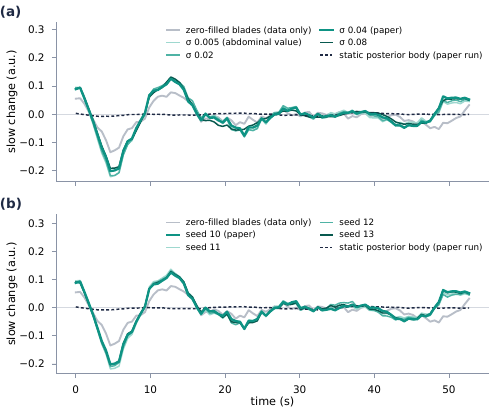}
    \caption{\textbf{Bowel motion of the \invivo{} scan across settings.} Slow intensity change (9- minus 31-frame moving average) at the first arrow of \cref{fig:teaser}c (5.2\,s), the strongest change along its line in the zero-filled data, over all 92 frames: (a) data-consistency weights $\sigma$, (b) four independent noise draws at $\sigma=0.04$. Gray: the zero-filled blades (data only; smaller, since a single blade blurs the image); dashed: the static posterior body. The traces correlate with the zero-filled one at 0.90--0.92 and with that of the GRAPPA blades (not shown) at 0.98 or more.}
    \label{fig:suppl-invivo-traces}
\end{figure}

\begin{figure}[t]
    \centering
    \includegraphics[width=\linewidth]{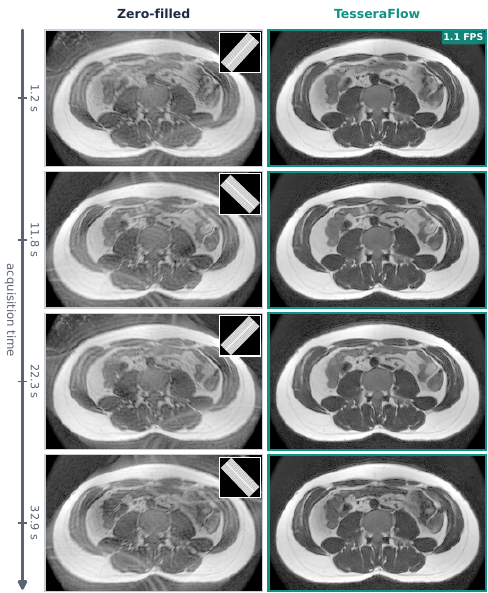}
    \caption{\textbf{Free-breathing \invivo{} PROPELLER scan}: frames spread over one 56-frame video, each from the single blade acquired at that time (inset). Left: gridding of the blade; right: \method{} (badge: reconstruction speed).}
    \label{fig:suppl-invivo-breathing}
\end{figure}
\begin{figure*}[t]
    \centering
    \includegraphics[width=\textwidth]{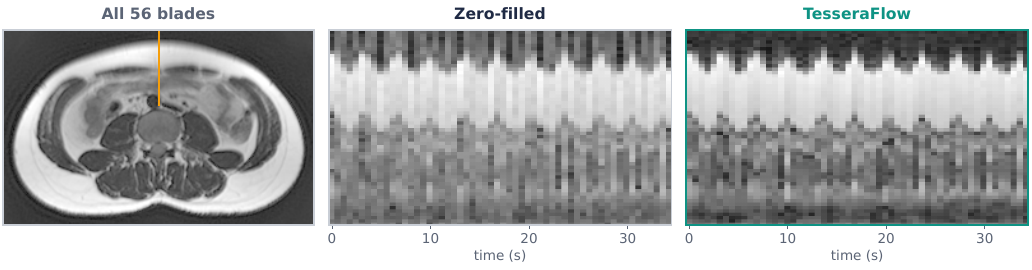}
    \caption{\textbf{x-t view of the free-breathing \invivo{} scan} of \cref{fig:suppl-invivo-breathing}: the centre column of the body across the anterior abdominal wall, which moves with breathing, marked on the conventional reconstruction, one image from all 56 blades (left), over all 56 frames, from the gridding of each frame's own blade (zero-filled, one display gain for all frames) and from \method{}.}
    \label{fig:suppl-invivo-breathing-xt}
\end{figure*}
\begin{figure}[t]
    \centering
    \includegraphics[width=\linewidth]{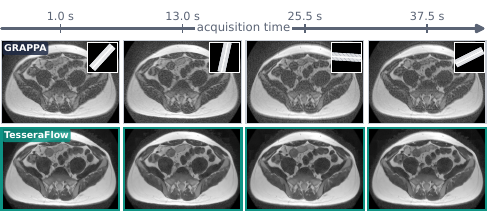}
    \caption{\textbf{Third \invivo{} PROPELLER scan} (breath hold, native $512\times512$): frames spread over one 78-frame video, each from the single blade acquired at that time (inset: acquired lines white, GRAPPA-filled lines gray). Top: the blade with GRAPPA; bottom: \method{}, with the model and settings of \cref{fig:invivo}.}
    \label{fig:suppl-invivo-exam2}
\end{figure}
\begin{figure*}[t]
    \centering
    \includegraphics[width=\textwidth]{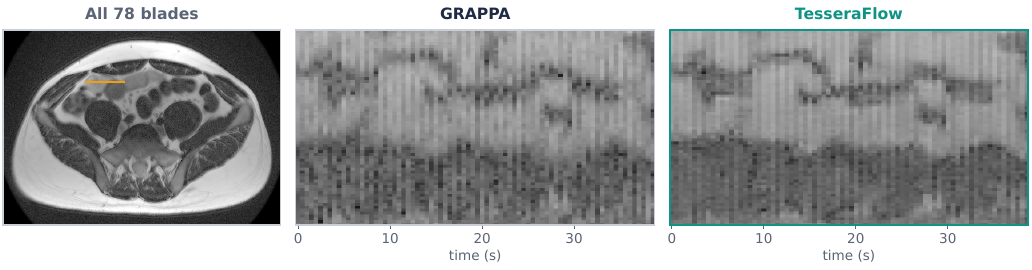}
    \caption{\textbf{x-t view of the third \invivo{} scan} of \cref{fig:suppl-invivo-exam2}: a line through bowel loops chosen from the blade data alone, marked on the conventional reconstruction, one image from all 78 blades (left), over all 78 frames, from each frame's own blade with GRAPPA (one display gain for all frames) and from \method{}.}
    \label{fig:suppl-invivo-exam2-xt}
\end{figure*}
\paragraph{Native resolution.} For \cref{fig:native}, the lung CT data are rebuilt at $512\times512$ with the same preprocessing and operator, and \method{} uses $\sigma=0.5$ and $t_s=0.6$ ($t_s=0.7$ and 0.8 lost 0.30 and 1.38\,dB on validation; 0.4--0.6 were added after this result). ADMM-TV (1{,}000 iterations) reaches 37.42\,dB and filtered backprojection 23.12\,dB. In training, \method{} on whole frames needs 46\,GB at $512^2$, and it and the whole-frame U-Net exceed a 96\,GB GPU at $1024^2$.

\paragraph{\Invivo{} PROPELLER.}
\cref{fig:invivo} applies the abdominal model to the raw data of a T2-weighted fast spin echo (FSE) PROPELLER scan from an exam outside our dataset, acquired from a volunteer during an attempted breath-hold, with written informed consent under a protocol approved by \irbname{}. It has an axial 3\,mm slice, repetition time 579\,ms, effective echo time 157\,ms, and 17 coils. One blade is acquired per echo train and repetition time, each rotated by the golden angle (111.25$^\circ$) from the previous one, so the 92 blades span 53.3\,s, 0.58\,s apart. We reconstruct on the scanner's native $512\times512$ grid (38\,cm field of view, 0.74\,mm pixels) with the abdominal model trained at $256\times256$, unchanged. At this grid, its $64\times64$ global view has an eighth of the frame's resolution instead of a quarter, and each $64\times64$ tile covers a quarter of the frame area it covers at $256\times256$. Each echo train gives one blade of 45 phase-encode lines with the scanner's in-blade acceleration ($R=3$): the 11 central lines and every third line over a span of 112. As in the scanner's own reconstruction, the skipped lines of each blade are filled from its central lines with a GRAPPA-type kernel before the standard PROPELLER phase correction, and the data consistency of each frame uses only its own filled blade. This filling correlates the noise of the filled lines. Using only the 45 acquired lines of each blade in $A$ instead gives the same bowel motion, with slow changes along the line of \cref{fig:teaser}c that correlate with those of the zero-filled blades at 0.79 ($\sigma=0.03$; 0.78 with the filled blades). Coil maps are estimated by ESPIRiT~\cite{uecker2014espirit} from all blades, and the image is normalized by the 99th percentile of the all-blade reconstruction. The sampler settings are those of the abdominal experiments except 120 steps (48 coarse steps before $t_s=0.8$ and 72 after) and $\sigma$, which we set to 0.04 by visual inspection of this scan; there is no reference for the moving anatomy. The conventional reconstruction combines all 92 blades into one image, as the scanner does, while \method{} resolves one frame per blade, as \cref{fig:suppl-invivo-xt} shows along a line through bowel loops on the other side of the abdomen from the line of \cref{fig:teaser}c. The anterior abdominal wall stays nearly still during the breath-hold; traced along the image column through the body's center, that of \method{} follows that of the single blades, which use the measurements alone, with correlation 0.92 for both the GRAPPA and the zero-filled blades (range 5.7 against 4.8 and 4.7\,mm). Fat edges also jitter by about half a pixel from frame to frame. This is fat--water chemical shift rather than motion. Fat is displaced along each blade's readout direction, which rotates with the blade, and consecutive blades therefore displace it in different directions. At the skin edge the displacement lies along the readout (0.57 pixels on average), whereas the posterior muscles show no consistent displacement, and the conventional reconstruction averages it out over all blades. Along the line of \cref{fig:teaser}c through the right-hand bowel loops, the slow intensity changes of \method{} (between 9- and 31-frame moving averages) correlate with those of the GRAPPA blades, which data consistency uses, at 0.95, and with those of the zero-filled blades, which keep only the acquired lines, at 0.78, about as high as the two blade reconstructions correlate with each other (0.81). They are 12 times stronger than in the static posterior body. These changes are bowel peristalsis. The first arrow marks their strongest change in the zero-filled data, and the other two mark wave-like motion of the lower bowel wall in \method{} at 36 and 44\,s. These slow changes persist at every tested data-consistency weight, from $\sigma=0.005$, the abdominal value, to 0.08, and across noise realizations (\cref{fig:suppl-invivo-traces}). Their correlation with those of the GRAPPA and zero-filled blades along the line stays at 0.90--0.98 and 0.74--0.80, and four samples with independent noise agree with each other at 0.97 (mean pairwise correlation of the slow changes along the line, 0.966--0.973). On the five abdominal test series of at least 100 frames, which have a fully sampled reference, we ran the same pipeline (one simulated blade per frame as in \cref{sec:data}, 120 steps, and a line chosen from the zero-filled blades alone). Its slow changes correlate with those of the reference at 0.75--0.90, against 0.65--0.85 for the zero-filled blades, at 0.95--1.15 times the reference's amplitude, and two samples agree at 0.90--0.97. In \cref{fig:teaser}c, at this grid \method{} takes 8.2\,s per frame, LPGC with its published sampler 61\,s, and PaDIS exceeds a 96\,GB GPU; LPGC keeps its hard data consistency, which has no noise level to match the $\sigma$ set for \method{}. \videoonly{The supplementary video \texttt{invivo\_peristalsis.mp4} plays each blade with GRAPPA, \method{}, and the conventional reconstruction frame by frame at three times real time, with a zoom on these bowel loops and \method{} along the line of \cref{fig:teaser}c.}

\paragraph{Held-out blades.} Without a reference, we test whether the reconstructed frames agree with measurements they did not use. Every eighth blade (11 of the 92, from blade 4 to blade 84) is withheld from the data consistency, so its frame is reconstructed from the prior and the neighboring frames alone. Each withheld blade is then predicted from its frame through the forward operator. With the settings of \cref{fig:teaser}c, the NRMSE between this prediction and the measured, filled blade is 0.408 for \method{} and 0.448 for LPGC, and lower for \method{} on all 11 blades. The mean of the two neighboring frames predicts the withheld blades with NRMSE 0.416 and 0.432. The frame of \method{} is thus closer to the measurements than this interpolation on all 11 blades, and that of LPGC on 2. Reconstructions from all blades fit the same blades to 0.261 and 0.207; LPGC fits them more closely because its hard data consistency also fits their noise. Each prediction is scaled by one real gain fitted on the frames that kept their data (1.02 and 1.00). The coil maps and the normalization still come from all blades, and withholding the 11 blades changes the normalization by 0.05\%.

\paragraph{Temporal response.} The prior couples five frames, so we measure how far a change spreads in time. In each of the 7 abdominal evaluation series, a 50\% brightness increase in a disk of radius 6 pixels in quiet tissue is added to the reference and to the coil k-space, either from frame 10 on or in frame 10 only, and \method{} (60 steps) reconstructs each series with and without it from the same seed. Averaged over the disk, the reconstruction recovers a median 89\% (80--92\%) of a lasting change in its first frame and 85\% (68--89\%) of a one-frame change in that frame, with a median 2\% (at most 9\%) in each neighboring frame and none further away. Two seeds of the unchanged series differ by 1\% (root mean square).

\paragraph{A second \invivo{} scan.} \cref{fig:suppl-invivo-breathing,fig:suppl-invivo-breathing-xt} apply the same abdominal model, trained at $256\times256$, to a free-breathing, T2-weighted FSE PROPELLER scan from a second exam outside our dataset, acquired from a volunteer with written informed consent under a protocol approved by \irbname{}. It has axial 3\,mm slices, repetition time 620\,ms, effective echo time 148\,ms, and 26 coils. Each slice is acquired in its own pass, one blade per echo train and repetition time, and the slice's 56 blades span 34.7\,s, 0.62\,s apart. We reconstruct on a $256\times256$ grid (36\,cm field of view, 1.4\,mm pixels). Each echo train gives one blade of 50 phase-encode lines, the 8 central lines and every second line over a span of 92, and the data consistency of each frame uses only its own blade, after the standard PROPELLER phase correction. Fat edges jitter between frames by about half a pixel along each blade's readout and not along its phase encode, consistent with the fat--water chemical shift described for the first scan; we leave it uncorrected. Coil maps and normalization are as for the first scan. The sampler settings are those of the abdominal experiments (60 steps, $t_s=0.8$) except $\sigma$, which we set to 0.02 by visual inspection of this scan, because with the abdominal value of 0.005 each frame fits its blade more closely and looks grainier. There is no reference for the moving anatomy. The conventional reconstruction combines all 56 blades into one image, collapsing 34.7\,s of breathing into one frame, while \method{} resolves one frame per blade. Along the image column through the body's center, the anterior abdominal wall moves with every breath (about 3.4\,s) in the single blades and in \method{}, which removes their streaks (\cref{fig:suppl-invivo-breathing-xt}).

\paragraph{A third \invivo{} scan.} \cref{fig:suppl-invivo-exam2,fig:suppl-invivo-exam2-xt} apply the model and settings of \cref{fig:invivo}, unchanged ($\sigma=0.04$; 120 steps, 48 before $t_s=0.8$ and 72 after), to a breath-hold, T2-weighted FSE PROPELLER scan from a third exam outside our dataset, acquired under the same protocol and consent. It has an axial 5\,mm slice at the level of the iliac wings, repetition time 500\,ms, effective echo time 121\,ms, and 26 coils, and its 78 blades span 39\,s, 0.5\,s apart. We reconstruct on the scanner's native $512\times512$ grid (38\,cm field of view, 0.74\,mm pixels). Each blade has 45 acquired phase-encode lines with in-blade acceleration $R=4$, and its skipped lines are filled with a GRAPPA-type kernel as for the first scan. Coil maps and normalization are also as for the first scan, and nothing was tuned on this scan. The line of \cref{fig:suppl-invivo-exam2-xt} is chosen from the blade data alone. Each filled blade is gridded alone and low-passed to a k-space disc that lies inside every blade, which removes the flicker with the blade angle, and the line passes through the largest slow change (between 5\,s and 18\,s moving averages) in the anterior half of the body, at least 50 pixels inside it, along the direction with the larger change. Bowel contents move along this line in the GRAPPA blades and in \method{}. \videoonly{The supplementary video \texttt{invivo\_exam2.mp4} plays this scan as the first one is played.}

\section{Potential negative impact}
\label{sec:suppl-impact}
Like other generative priors, \method{} can produce plausible but incorrect structure where the measurements constrain the image weakly, and in a video such errors could be mistaken for motion. Our quantitative evaluation undersamples fully sampled data retrospectively and scores image quality, and the \invivo{} reconstruction has no reference for the moving anatomy. The reconstructions are therefore not validated for diagnosis, which would require prospective studies with radiologist readers.

\let\setcounter\arxivsetcounter

\end{document}